\documentclass{article}

\PassOptionsToPackage{numbers,compress}{natbib}
\usepackage[preprint]{neurips_2026}

\usepackage[utf8]{inputenc} 
\usepackage[T1]{fontenc}    
\usepackage{hyperref}       
\usepackage{url}            
\usepackage{booktabs}       
\usepackage{amsfonts}       
\usepackage{nicefrac}       
\usepackage{microtype}      
\usepackage{xcolor}         
\usepackage{amsmath}
\usepackage{amssymb}
\usepackage{mathtools}
\usepackage{algorithm}
\usepackage{algorithmic}
\usepackage{amsthm}
\usepackage{enumitem}
\usepackage{multirow}
\usepackage{graphicx}
\usepackage[capitalize,noabbrev]{cleveref}
\usepackage{bm}
\theoremstyle{plain}
\newtheorem{theorem}{Theorem}[section]

\newtheorem{lemma}[theorem]{Lemma}

\theoremstyle{definition}

\theoremstyle{remark}

\title{Efficient Decentralized Multi-task Dataset Valuation via Model Merging}

\author{
Mohammadsajad Alipour \\
Department of Computer Science\\
Rensselaer Polytechnic Institute\\
Troy, NY 12180, USA\\
\texttt{alipom@rpi.edu}
\And
Mohammad Mohammadi Amiri \\
Department of Computer Science\\
Rensselaer Polytechnic Institute\\
Troy, NY 12180, USA\\
\texttt{mamiri@rpi.edu}
}

\begin{document}

\maketitle

\begin{abstract}
  Accurate and efficient dataset valuation is essential for enabling fair and transparent data marketplaces, especially when multiple contributors provide data for training multi-task models.\ Most existing valuation methods, however, are limited to single-task settings, overlooking scenarios where a buyer aims to optimize performance across multiple downstream tasks. Moreover, traditional valuation approaches—such as Shapley-based or retraining-based methods—are computationally expensive and poorly suited for decentralized environments without a trusted central coordinator and with strict privacy constraints. We propose DMVM (Decentralized Multi-task Valuation via Model Merging), a novel framework that bypasses retraining and data sharing by leveraging task arithmetic to infer dataset contributions directly from model combinations. Instead of retraining or sharing raw data, DMVM quantifies how models trained on different datasets combine in parameter space to infer each dataset’s marginal utility across multiple tasks. This formulation yields a valuation process that is scalable, computationally efficient, and explicitly aligned with multi-task generalization behavior. To support decentralized deployment, we introduce a secure aggregation protocol that enables collaborative valuation without revealing individual model parameters or private data. We also provide theoretical error bounds characterizing the approximation quality of DMVM and validate our framework through comprehensive experiments on computer vision and natural language processing tasks. 
\end{abstract}

\section{Introduction}
Dataset valuation has emerged as a central challenge in the design of modern data marketplaces, where multiple sellers offer proprietary datasets and buyers seek to improve model performance on specific target tasks. Accurately estimating the value of a seller’s data prior to purchase is vital for fair pricing, informed decision-making, and sustaining trust in the marketplace. This estimation aims to quantify the marginal utility a dataset provides to a learning algorithm for a specific task, guiding buyers toward high-impact acquisitions and incentivizing high-quality data contributions from sellers.

Prior work has made substantial progress toward formalizing the problem of data valuation. Game-theoretic methods such as Data Shapley \citep{pmlr-v97-ghorbani19c} offer theoretically principled metrics by averaging marginal contributions across all data subsets. Influence-function techniques estimate the effect of training samples on model predictions \citep{koh2017understanding,zhang2025toward}, while gradient-based methods (e.g., reinforcement learning or utility-based proxies) seek to capture a dataset’s alignment with target tasks \citep{yoon2020data,cai2024chg}. While theoretically compelling, these methods face critical obstacles when applied to real-world data marketplaces, as discussed next.
A primary barrier to practical deployment is computational overhead: estimating marginal contributions, especially via Shapley-based methods, typically requires retraining or fine-tuning models across an exponential number of dataset combinations. This becomes infeasible at scale, particularly for large models or a large number of sellers. A second challenge is data privacy. Most valuation techniques assume centralized access to raw data, an unrealistic expectation in many real-world marketplaces involving sensitive or regulated datasets, e.g., medical imaging or personal data. In such settings, sellers are unlikely to permit raw data sharing, even for valuation purposes. 

Beyond scalability and privacy, a third and often overlooked limitation is the narrow focus on single-task valuation. Most existing methods assess a dataset's utility with respect to a single predefined task. However, in realistic settings, buyers often pursue multi-task learning objectives (e.g., multi-task vision models performing both detection and segmentation, or multi-task language models performing sentiment analysis, topic classification, and summarization). Valuation in a multi-task learning context is inherently more complex because a dataset’s utility may differ across tasks, and its value cannot be captured by a single performance metric. Despite the growing importance of multi-task models in practice, there has been little work addressing dataset valuation in this setting, leaving a significant methodological gap \cite{ahmadian2024mix,zhu2026multi}.

Meanwhile, model merging has recently emerged as a strategy for combining models fine-tuned on different datasets without requiring access to, or retraining on, combined data \citep{ilharco2023editing,wortsman2022model,matena2022merging,wang2025scaling,alipour2025towards}. Starting from a common base, models are fine-tuned independently; their parameter updates, often represented as task vectors \citep{ilharco2023editing} are then merged—via averaging, interpolation, or more advanced techniques such as task-vector trimming or balancing—to approximate the effect of joint training \citep{yadav2023ties,yu2024language}. Empirically, \citet{tao2025merge} showed that merging models fine-tuned on disjoint datasets can closely approximate the performance of a model trained on the combined data. They further show a strong linear correlation between the accuracy of the merged model and that of the model trained on the full data mixture, indicating that model merging can serve as an efficient and reliable proxy for mixture fine-tuning or multi-task learning. Moreover, \citet{zhou2025task} provide theoretical support for this intuition by deriving an error bound between the weights of a model trained via multi-task learning and a model merged with task-arithmetic.

In multi-task setting, the ideal way to assess a dataset’s value is to measure its marginal improvement to a jointly trained multi-task model; that is, how much performance gain it contributes when all tasks are optimized jointly. However, such joint training is computationally prohibitive in decentralized marketplaces, as it requires access to all datasets and repeated retraining across combinations of sellers. To overcome this bottleneck, we draw inspiration from recent advances in model merging, which approximate the effects of joint multi-task training without the cost of training on combined datasets. Starting from a shared initialization, models are fine-tuned independently on individual datasets, and their parameter updates, often represented as task vectors, are combined through arithmetic operations in parameter space. Empirical and theoretical studies have shown that this process can closely replicate the outcomes of true multi-task learning \cite{li2025when,ilharco2023editing,zhou2025task,na2024scalable}, suggesting that model merging can serve as an efficient proxy for evaluating cross-task contributions.

Building on this insight, we introduce DMVM (Decentralized Multi-task Valuation via Model Merging), a framework that formalizes multi-task dataset valuation through the lens of model merging. DMVM estimates each dataset’s marginal utility by analyzing how its corresponding model influences the aggregated performance across a buyer’s target tasks. Our formulation enables fully decentralized valuation without raw data exchange, incorporates a privacy-aware protocol to protect individual models during valuation, and provides a theoretical error bound that characterizes the approximation gap between DMVM-based valuation and the ground-truth multi-task valuation. We validate our framework through simulations of a decentralized data marketplace, showing consistent value estimates across vision and language benchmarks.
Our main contributions are:
\begin{itemize}
[leftmargin=*,noitemsep,topsep=0pt]
    \item \textbf{Efficient and decentralized valuation framework.} We introduce DMVM, a novel approach for multi-task dataset valuation through model merging. It operates without retraining or centralized data access, enabling a fully decentralized deployment, without a trusted broker or third party. 
    \item \textbf{Privacy-aware protocol.} DMVM incorporates a secure aggregation protocol enabling collaborative valuation without revealing raw data or individual models, ensuring compatibility with privacy-sensitive applications. 
    \item \textbf{Theoretical guarantees.} We derive error bounds that characterize the approximation quality of DMVM relative to ground-truth multi-task marginal contributions.
    \item \textbf{Empirical evaluation.} We demonstrate stable and reliable valuation performance of DMVM across representative vision and language benchmarks.
\end{itemize}
By bridging the domains of dataset valuation and model merging, DMVM establishes a scalable, privacy-aware, and theoretically grounded foundation for dataset valuation in multi-task marketplaces.
\section{Related Work}
The proliferation of data-driven machine learning has created a need to quantify the economic value of data, both to fairly compensate contributors and to efficiently acquire new, high-quality data. This research area can be broadly categorized into methods for valuing individual data points, techniques for valuing entire datasets, and the design and operation of data marketplaces.

\textbf{Data Valuation: Valuing Individual Points.}
A dominant approach for data valuation is the Shapley value \citep{shapley1953value}, a concept from cooperative game theory that fairly assigns value to each data point based on its average marginal contribution to a machine learning model's performance across all possible data subsets \citep{pmlr-v97-ghorbani19c}. This method is axiomatically fair, satisfying properties such as symmetry and additivity.
The primary drawback of the standard Data Shapley is its exponential computational complexity, $O(2^N)$, which is prohibitive for modern-scale datasets of size $N$. To overcome this, several approximations have been developed \citep{pmlr-v89-jia19a}.
Sampling-based methods estimate the Shapley value by sampling random permutations of the data, but they may still require numerous model retrainings \citep{pmlr-v97-ghorbani19c,maleki2013bounding}. \citet{knnshap} proposed an exact and efficient algorithm for computing Shapley values specifically for K-nearest neighbour models, reducing the complexity to $O(N\log{N})$.
Recent works further extend Shapley-based valuation to graph-structured data \citep{chi2025precedenceconstrained,11150137graph}.

\textbf{Dataset Valuation.}
While point-level valuation supports curation and debugging, dataset valuation addresses how much an entire contributor’s dataset adds to utility within a cooperative game among data owners. This line of work focuses on valuing entire datasets as a single unit, which is distinct from aggregating individual data point values. The Shapley value can also be applied at the dataset level, measuring the marginal contribution of one seller's entire dataset to a coalition of other sellers. To address the computational cost, DU-Shapley \citep{garrido2024shapley} was proposed as an efficient proxy specifically for dataset valuation, using a discrete uniform approximation that is more efficient than generic Monte Carlo sampling.
Another line of work is task-agnostic dataset valuation methods that assess data without relying on a specific model or validation set, avoiding issues such as unavailable validation data. Instead, they estimate the value based on the dataset’s intrinsic distributional properties. The PriArTa framework evaluates sellers by computing the Wasserstein distance between augmentation-invariant latent distributions of the buyer's and seller's datasets  \cite{jahani2024private}. Another related study introduces a fundamental valuation framework that characterizes datasets through two complementary metrics: relevance, capturing statistical similarity, and diversity, capturing statistical difference \citep{amiri2023fundamentals}. These are estimated by comparing the dataset's second-moment properties, which can be done privately by sharing only principal components. This methodology was later extended to graph data \cite{falahati2024disentangled}.

\textbf{Data Marketplace.}
A growing line of research explores the design and economic principles of data marketplaces. Building such markets is challenging because data possess unique characteristics: they can be replicated at near-zero cost, their value arises from complex interactions among datasets, and their usefulness for a given task is difficult to evaluate beforehand \citep{dealer,tian2022private,chhachhi2024joint,modelbased,algorith,lu2024data}.
To address this, some frameworks propose a model-based pricing approach, where the marketplace sells machine learning model instances rather than the raw data itself \citep{modelbased}. The Dealer marketplace framework \citep{dealer} formalizes an end-to-end system where data owners' compensation is a function of both their Shapley value and their privacy sensitivity. These mechanisms, however, often rely on a trusted broker to access all datasets and perform the valuation. This introduces a fundamental dilemma: data owners must share their data to have it valued, but a buyer no longer needs to purchase data they have already observed. Privatizing these computations via multi-party computation (MPC)-style protocols closes the loop for deployable and trustable data markets \citep{tian2022private,ouyang2025selectformer}. 

\textbf{Model Merging.}
Model merging builds on the observation that neural network parameters often exhibit approximately linear structure across tasks. \citet{ilharco2023editing} first introduced task arithmetic, showing that subtracting a pre-trained model from its fine-tuned counterpart yields a task vector whose addition or subtraction can edit model behavior, enabling multi-task composition or unlearning without retraining.
\citet{li2025when} provided the first theoretical justification for task arithmetic, proving that the effectiveness of adding or subtracting task vectors depends on how related the underlying tasks are and showing that properly scaled combinations can generalize to unseen tasks. In an empirical study, \citet{tao2025merge} proposed Merge to Mix, which leverages merging as a surrogate for fine-tuning on dataset mixtures. By correlating the performance of merged models with that of mixture-trained ones, they enabled efficient and data-free selection of optimal dataset combinations. 
\citet{zhou2025task} reframed task-arithmetic as an efficient form of approximate multi-task learning. They showed that merging task vectors is equivalent to performing one step of gradient descent on a combined multi-task objective, and derived a second-order bound quantifying the difference between the merged model and the model obtained from full multi-task training. This analysis formally explains why task arithmetic can closely approximate joint optimization.
Despite these advances, existing model merging studies have not been explored in the context of dataset valuation, nor leveraged to quantify marginal data contributions in decentralized multi-task settings.
\begin{figure}[t]
    \centering
\includegraphics[width=0.5\linewidth]{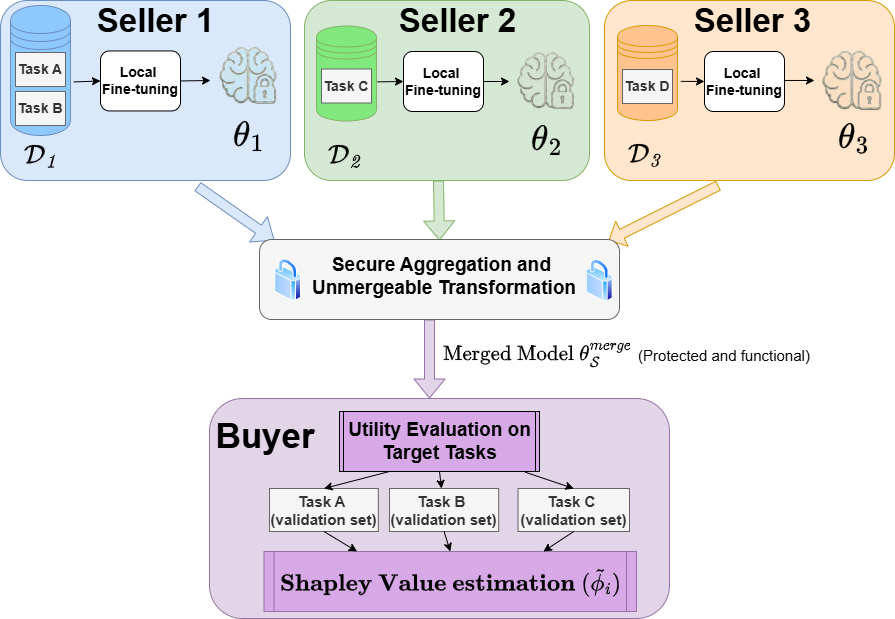}
    \caption{Overview of the DMVM framework for decentralized multi-task dataset valuation. Each seller trains a local model on its private dataset, and using a secure protocol, they produce a protected merged model for each coalition. The buyer evaluates the merged models on its target validation tasks to estimate seller contributions via Shapley values, without accessing sellers' private data or individual models.}
    \label{fig:Overview}
\end{figure}
\section{Preliminaries}

In a data marketplace, multiple sellers offer datasets of varying relevance and quality, and a buyer seeks to assess the contribution of each dataset to the performance of a learned model. This process, known as \emph{dataset valuation}, provides a quantitative basis for fair pricing and informed data acquisition. A principled valuation metric should capture the \emph{marginal utility} of each dataset; that is, how much additional benefit it brings when combined with others.
We consider a data marketplace consisting of a set of $n$ data sellers, where each seller $i$ holds a private dataset $\mathcal{D}_i$, and the collection of all datasets is denoted by $\mathcal{D}=\{\mathcal{D}_1,\dots,\mathcal{D}_n\}$. These datasets are not accessible to the buyer prior to purchase due to privacy or commercial constraints. The buyer is interested in a set of downstream tasks $\mathcal{T}$, where each task $t \in \mathcal{T}$ has a validation set $\mathcal{D}^{\text{val}}_{t}$. The buyer's objective is to train a model $\bm{\theta}$ that generalizes well across all tasks in $\mathcal{T}$.

If the buyer had access to the datasets of a subset of sellers $\mathcal{S} \subseteq [n]:=\{1,...,n\}$, a model could be trained via multi-task learning (MTL) as $
    \bm{\theta}^{\text{MTL}}_{\mathcal{S}}
    = \arg\min_{\bm{\theta}} \sum_{i \in \mathcal{S}} \mathcal{L}({\bm{\theta}}, \mathcal{D}_i),
$
where $\mathcal{L}$ denotes the loss function. The corresponding utility of the subset $\mathcal{S}$ is defined as the average performance of the resultant model across all downstream tasks:
\begin{align}\label{gt-utility}
    u(\mathcal{S}) = 
    \frac{1}{|\mathcal{T}|} 
    \sum_{t \in \mathcal{T}} 
    \mathrm{Perf}_{t}(\bm{\theta}^{\text{MTL}}_{\mathcal{S}}, \mathcal{D}^{\text{val}}_{t}),
\end{align}
where $\mathrm{Perf}_{t}$ denotes a task-specific evaluation metric (e.g., accuracy or negative loss) on task $t$. Alternatively, utility can be expressed in terms of validation loss as $
u(\mathcal{S}) = -\sum_{t \in \mathcal{T}} \mathcal{L}(\bm{\theta}^{\text{MTL}}_{\mathcal{S}}, \mathcal{D}^{\text{val}}_{t}).
$
The marginal contribution of seller $i$ to a coalition $\mathcal{S}$ of other sellers is given by $ u(\mathcal{S} \cup \{i\}) - u(\mathcal{S}),$
which quantifies the added utility from including dataset $\mathcal{D}_i$ when training jointly with datasets of the sellers in $\mathcal{S}$. 
The Shapley value~\citep{shapley1953value} provides a unique and principled method for aggregating marginal contributions across all possible coalitions. For seller $i$, it is defined as:
\begin{align}\label{gt-equ}
    \phi_i = 
    \frac{1}{n}
    \sum_{\mathcal{S}\subseteq[n]\setminus\{i\}} 
    \frac{u(\mathcal{S} \cup \{i\}) - u(\mathcal{S})}{\binom{n-1}{|\mathcal{S}|}},
\end{align}
where the summation averages seller $i$'s marginal contribution across all subsets $\mathcal{S}$ that excludes $i$, weighted by their cardinality. This formulation satisfies key fairness axioms, including symmetry, linearity, and additivity, and is therefore widely regarded as the \emph{ground-truth reference} for data valuation \cite{pmlr-v97-ghorbani19c}.
Despite its strong theoretical foundations, computing the Shapley value $\phi_i$ directly is computationally intractable in practice. It requires training models on all $2^{n-1}$ possible subsets $\mathcal{S}\subseteq[n]\backslash\{i\}$ to evaluate $u(\mathcal{S})$, which becomes exponentially expensive as $n$ grows. Moreover, in real marketplaces, buyers do not have access to the raw datasets of sellers prior to purchase. Therefore, the utility function $u(\mathcal{S})$ cannot be computed without violating confidentiality.
These challenges motivate the development of scalable and privacy-aware alternatives to the classical Shapley formulation. In the following section, we propose DMVM, a framework designed to overcome both computational and privacy limitations while maintaining theoretical faithfulness to multi-task marginal contributions.

\section{Methodology}

This section introduces our proposed framework for decentralized dataset valuation. 
Section~\ref{proxy} presents an efficient proxy for approximating the utility function $u(\cdot)$ without requiring model training. Section~\ref{priv} then introduces a privacy-aware protocol that enables secure valuation without exposing individual models.

\subsection{Model Merging as a Proxy for Utility}
\label{proxy}

We assume that all sellers either adopt a common pretrained model $\bm{\theta}^{\text{pre}}$ provided by the buyer or agree on a shared architecture. 
Each seller $i$ locally fine-tunes this model on their private dataset $\mathcal{D}_i$, yielding a personalized model $\bm{\theta}_i$.
To estimate the utility of any seller subset $\mathcal{S} \subseteq [n]$, 
the buyer merges the corresponding fine-tuned models $\{\bm{\theta}_i : i \in \mathcal{S}\}$ into a single model $\bm{\theta}^{\text{merge}}_\mathcal{S}$ using a merging strategy, specifically, \emph{task arithmetic}. 
For brevity, we use $\bm{\theta}$ to denote full model parameters and their associated weight matrices.
Building on recent theoretical and empirical findings~\citep{tao2025merge,zhou2025task}, 
we approximate the true utility $u(\mathcal{S})$ with the proxy $\hat{u}(\cdot)$ defined as $
    \hat{u}(\mathcal{S}) 
    = \frac{1}{|\mathcal{T}|} 
    \sum_{t \in \mathcal{T}} 
    \mathrm{Perf}_{t}\!\left(
    \bm{\theta}^{\text{merge}}_\mathcal{S}, 
    \mathcal{D}^{\text{val}}_{t}
    \right),$
or equivalently, in terms of loss as $
    \hat{u}(\mathcal{S}) = -\sum_{t \in \mathcal{T}} \mathcal{L}(\bm{\theta}^{\text{merge}}_{\mathcal{S}}, \mathcal{D}^{\text{val}}_{t}).
$
The merged model $\bm{\theta}^{\text{merge}}_\mathcal{S}$ is obtained as $
    \bm{\theta}^{\text{merge}}_\mathcal{S}
    = \bm{\theta}^{\text{pre}} + \alpha \sum_{i \in \mathcal{S}} (\bm{\theta}_i - \bm{\theta}^{\text{pre}}) ,
$
where $\alpha$ is a scaling hyperparameter \citep{ilharco2023editing} and $(\bm{\theta}_i - \bm{\theta}^{\text{pre}})$ denotes the task vector associated with seller $i$. 
Using this utility proxy, we approximate the Shapley value $\phi_i$ of each seller $i$ as:
\begin{align*}
    \hat{\phi}_i
    = \frac{1}{n}
    \sum_{\mathcal{S} \subseteq [n] \setminus \{i\}}
    \frac{
        \hat{u}(\mathcal{S} \cup \{i\})
        - \hat{u}(\mathcal{S})
    }{
        \binom{n-1}{|\mathcal{S}|}
    }.
\end{align*}

This formulation enables the buyer to estimate each dataset’s contribution without retraining a model for all $2^{n-1}$ possible subsets. 
By approximating utility $u(\cdot)$ with $\hat{u}(\cdot)$, the computational burden is reduced from $2^{n}$ model trainings and inferences to $2^{n}$ model summations and evaluations, yielding substantial efficiency gains while preserving a close approximation to the ground-truth Shapley value, as established in the following theorem:
\begin{theorem}
\label{thm:error_bound}
Assume that MTL is performed via full-batch gradient descent, 
and that the difference between the merged model $\bm{\theta}^{\text{merge}}_\mathcal{S}$ and the MTL-trained model $\bm{\theta}^{\text{MTL}}_\mathcal{S}$ is approximated up to second-order terms 
(i.e., higher-order components are negligible, with remainder dominated by $\mathcal{O}(\eta^3)$, where $\eta$ is the learning rate for MTL). 
If the loss function $\mathcal{L}$ is $L$-Lipschitz continuous, 
then the approximation error between the true and proxy Shapley values satisfies:
\begin{align*}
    |\phi_i - \hat{\phi}_i|
    \le
    (2\alpha n^2 - 2\alpha n + 2n + \alpha - 1)\, L C,
\end{align*}
where 
$C = \binom{h+2}{2} H_{\max} G_{\max} $. Here,
$h$ denotes the number of optimization steps (i.e., epochs), 
and $G_{\max}$ and $H_{\max}$ are upper bounds on the gradient and Hessian norms, respectively.
\end{theorem}

This theorem establishes a formal link between MTL and task-arithmetic as the model merging method, in the context of Shapley-based valuation (the proof is provided in Appendix~\ref{proofthm1}).
It shows that, under mild assumptions, the error between DMVM and ground-truth valuation, grows quadratically in the number of sellers and linearly in $\alpha$. Moreover, setting $\alpha=\frac{1}{n}$ makes the error grow linearly with $n$. However, this comes at the cost of restricting the performance of the merging algorithm, since $\alpha$ is typically treated as a tunable hyperparameter, e.g., selected via binary search.
Furthermore, even the $2^{n}$ summations and evaluations can be replaced by an arbitrary finite number of Monte Carlo (MC) samples while retaining a controlled approximation error, as established in the following theorem:
\begin{theorem}
\label{thm:mc_error_bound}
Let $\bar{\phi}_i^{(m)}$ denote the MC estimator of $\hat{\phi}_i$
obtained from $m$ independent samples of the marginal contribution terms in the
Shapley expansion. Assume each sampled marginal contribution lies in an interval
of width at most $R$. Under the assumptions of Theorem~\ref{thm:error_bound},
for any $\delta \in (0,1)$, with probability at least $1-\delta$, we have: 
\begin{align*}
    |\phi_i - \bar{\phi}_i^{(m)}|
\le
(2\alpha n^2 - 2\alpha n + 2n + \alpha - 1)\,LC
+
\sqrt{\frac{R^2}{2m}\log\!\left(\frac{2}{\delta}\right)}.
\end{align*}

\end{theorem}
Theorem~\ref{thm:mc_error_bound} shows that DMVM can admit a further sampling-based
acceleration beyond the deterministic approximation in Theorem~\ref{thm:error_bound}.
In particular, the total error decomposes into a deterministic merging error and a
statistical sampling error that decays as $\mathcal{O}(1/\sqrt{m})$. Proof for this theorem is provided in Appendix~\ref{montecarl}.
Finally, although the utility proxy $\hat{u}(\cdot)$ avoids raw data sharing, computing the merged model $\bm{\theta}^{\text{merge}}_\mathcal{S}$ still requires access to the individual fine-tuned models $\{\bm{\theta}_i\}_{i\in[n]}$. To prevent model leakage during this process, we introduce a privacy-aware protocol in the next section.

\subsection{Privacy-Aware Merging via Secure Aggregation}\label{priv}

While the utility proxy $\hat{u}(\cdot)$ enables efficient valuation, it still requires access to the individual fine-tuned models $\{\bm{\theta}_i\}_{i\in [n]}$ for computing the merged model $\bm{\theta}^{\text{merge}}_\mathcal{S}$. Directly sharing these models would violate seller privacy and undermine the decentralized design of the marketplace. To address this, we design a \emph{two-fold privacy-aware protocol} that integrates secure MPC with an \emph{unmergeable transformation} mechanism inspired by recent advances in model obfuscation~\citep{Junhao_2025_ICCV,wang2025model,chen2025defending}.

\textbf{Overview.}
Our protocol incorporates two components:  
(1)~a \emph{secure aggregation} mechanism that allows the buyer to obtain only the aggregated merged model $\bm{\theta}^{\text{merge}}_\mathcal{S}$ without revealing any individual model $\bm{\theta}_i$; and  
(2)~an \emph{unmergeable transformation} step that prevents an adversarial buyer from reconstructing any seller’s model via algebraic manipulation of overlapping subsets of the merged models.  
We first describe the secure aggregation process, then extend it to incorporate the unmergeable mechanism.

\textbf{Secure Aggregation.}
For brevity and simplicity, we set $\alpha=\frac{1}{|\mathcal{S}|}$; we have provided the algorithm for any arbitrary $\alpha$ value in Appendix~\ref{mainalg}. For any subset $\mathcal{S}\subseteq[n]$ with $|\mathcal{S}| \ge 2$, each seller $i \in \mathcal{S}$ samples a random mask matrix $\bm{m}_i$ of the same dimension as their model parameters.  
Each seller shares ${\bm{m}}_i$ with all other sellers in $\mathcal{S}$ and computes a masked version of their contribution as
$
    \bm{e}_i
    = 
    \frac{\bm{\theta}_i}{|\mathcal{S}|}
    - {\bm{m}}_i,
$
which is sent to the buyer. Meanwhile, all sellers compute the aggregate mask ${\bm{m}} = \sum_{i \in \mathcal{S}} {\bm{m}}_i$ and send it to the buyer, who verifies the consistency across submissions.
The buyer then reconstructs the merged model as $
    \bm{\theta}^{\text{merge}}_\mathcal{S} 
    = {\bm{m}} + \sum_{i \in \mathcal{S}} \bm{e}_i = \sum_{i \in \mathcal{S}} {\bm{m}}_i + 
       \sum_{i \in \mathcal{S}}\!\left(
      \frac{\bm{\theta}_i}{|\mathcal{S}|}
    - {\bm{m}}_i
       \right) = \frac{1}{|\mathcal{S}|} \!\sum_{i \in \mathcal{S}} \bm{\theta}_i.
$
This guarantees that the buyer receives only the aggregated model $\bm{\theta}^{\text{merge}}_\mathcal{S}$, not individual $\bm{\theta}_i$'s.  
Similarly, since only random masks are shared among sellers, no seller gains access to others' models.

\textbf{Unmergeable transformation.}
While the above protocol protects individual models, a malicious buyer could still recover a seller’s model by subtracting two merged models from overlapping subsets (e.g., $\{\theta_1, \theta_2, \theta_3\}$ and $\{\theta_1, \theta_2\}$).  
To prevent this, we introduce a transformation that preserves functional behavior for inference but disrupts linear relationships among models. We instantiate this for two common modules: multi-layer perceptron (MLP) blocks and consecutive linear projections.

Consider two consecutive layers of an MLP:
$\text{MLP}(x) = \bm{W}^{(2)} \sigma(\bm{W}^{(1)} \bm{x} + \bm{b}^{(1)}) + \bm{b}^{(2)}$, where $\sigma(\cdot)$ is an element-wise nonlinearity. For weight matrices $\bm{W}^{(1)}_i$ and $\bm{W}^{(2)}_i$ held by seller $i$, the goal is for the buyer to recover $\sum_{i\in\mathcal{S}} \bm{W}^{(1)}_i$ and $\sum_{i\in\mathcal{S}}\bm{W}^{(2)}_i$ without accessing individual weights. To achieve this, all sellers in $\mathcal{S}$ jointly agree on a random permutation matrix $\bm{P}_\mathcal{S}$ (orthogonal with one “1” per row/column and “0” elsewhere, satisfying $\bm{P}_\mathcal{S}^{-1} = \bm{P}_\mathcal{S}^\top$).  
For mask matrices ${\bm{m}}^{(1)}_i$ and ${\bm{m}}^{(2)}_i$, instead of sending $\frac{\bm{W}^{(1)}_i}{|\mathcal{S}|} - {\bm{m}}^{(1)}_i$ and $\frac{\bm{W}^{(2)}_i}{|\mathcal{S}|} - {\bm{m}}^{(2)}_i$, 
each seller $i$ transmits:
\begin{align*}
    \frac{\bm{P}_\mathcal{S}\bm{W}^{(1)}_i}{{|\mathcal{S}|}} - {\bm{m}}^{(1)}_i ,\quad 
\frac{\bm{W}^{(2)}_i \bm{P}_\mathcal{S}^\top}{{|\mathcal{S}|}} - {\bm{m}}^{(2)}_i.
\end{align*}

Having received $\sum_{i\in\mathcal{S}}\bm{m}^{(1)}_i$ and $\sum_{i\in\mathcal{S}}\bm{m}^{(2)}_i$ from the sellers, the buyer can only reconstruct $\frac{\sum_{i\in\mathcal{S}}\bm{P}_\mathcal{S}\bm{W}^{(1)}_i}{|{\mathcal{S}}|}$ and $\frac{\sum_{i\in\mathcal{S}}\bm{W}^{(2)}_i\bm{P}_\mathcal{S}^\top}{|{\mathcal{S}}|}$ instead of $\frac{\sum_{i\in\mathcal{S}}\bm{W}^{(1)}_i}{|{\mathcal{S}}|}$ and $\frac{\sum_{i\in\mathcal{S}}\bm{W}^{(2)}_i}{|{\mathcal{S}}|}$. Since $\bm{P}_\mathcal{S}^\top \bm{P}_\mathcal{S}=I$ and $\sigma(\cdot)$ operates elementwise, this transformation leaves the model's inference behavior unchanged, but destroys the additive structure that would otherwise enable model reconstruction via subtraction.
Thus, any negation or subtraction of merged models yields uninformative parameter combinations.
For consecutive linear layers such as $\bm{QK}^\top$ projection in transformer blocks~\citep{vaswani2017attention} (for every matrix $\bm{Q}$ and key matrix $\bm{K}$), 
a similar mechanism applies. Sellers agree on a random scalar $\rho \in \mathbb{R}\backslash\{0\}$, and instead of sending $\frac{\bm{Q}_i}{|\mathcal{S}|}-\bm{m}^{(1)}_i$ and $\frac{\bm{K}_i}{|\mathcal{S}|}-\bm{m}^{(2)}_i$, seller $i$ sends $\frac{\rho \bm{Q}_i}{|\mathcal{S}|}-\bm{m}^{(1)}_i$ and $\frac{\bm{K}_i}{\rho|\mathcal{S}|}-\bm{m}^{(2)}_i$ to the buyer.
This also preserves the functionality of $\bm{QK}^\top$ while making the individual components unmergeable, and therefore negations would result in useless and meaningless weights.

\begin{table}
\centering
\scriptsize
\caption{Comparison of our methods and baselines ($m$ is the number of subsets that are sampled).}
\label{tab:methods_comparison}
\begin{tabular}{l c c}
\hline
\textbf{Method} & \textbf{Operates without access to sellers' datasets} & \textbf{Computational time} \\
\hline
Dataset Shapley (GT) & $\times$ & $2^n\times (T_{\text{train}}+T_{\text{inference}})$ \\
MC-Dataset-Shapley \citep{pmlr-v97-ghorbani19c} & $\times$ & $n\times {m}\times (T_{\text{train}}+T_{\text{inference}})$ \\
DU-Shapley \citep{garrido2024shapley} & $\times$ & $n\times{n}\times (T_{\text{train}}+T_{\text{inference}})$ \\
\midrule
\textbf{DMVM} & \checkmark & $2^{n}\times T_{\text{inference}}$  \\
\textbf{MC-DMVM} & \checkmark & $n\times{m}\times T_{\text{inference}}$ \\
\hline
\end{tabular}

\end{table}
\textbf{Secure Shapley Estimation.}
With these mechanisms in place, the buyer never accesses any raw data or individual model weights. We define the privacy-aware Shapley estimate of seller $i$ as:
\begin{align}\label{eq:pri-shap}
    \tilde{\phi}_i=\frac{1}{n}\sum_{\mathcal{S}\subseteq [n]\backslash\{i\},|\mathcal{S}|\geq2} \frac{\hat{u}(\mathcal{S}\cup\{i\})-\hat{u}(\mathcal{S})}{\binom{n-1}{|\mathcal{S}|}}.
\end{align}
where $\hat{u}(\cdot)$ is computed via the above secure aggregation mechanism with unmergeable transformation. In the above approximation, we restrict the cardinality to $|\mathcal{S}|\geq2$, thereby excluding the case $|\mathcal{S}|=1$ to guarantee that no individual model is shared from a seller to the buyer, which is required for privacy. The following results bound the deviation between $\tilde{\phi}_i$ and the true Shapley value $\phi_i$ with the proof provided in Appendix~\ref{proofthm2}.

\begin{theorem}
Let $R$ denote the maximum marginal contribution $|u(\mathcal{S} \cup \{i\}) - u(\mathcal{S})|$ over all subsets in the Shapley expansion of $\phi_i$. Then,
\begin{align*}
    |\phi_i - \tilde{\phi}_i|
    \le
    \frac{n - 2}{n}(2\alpha n^2 - 2\alpha n + 2n + \alpha - 1)\,LC
    + \frac{2R}{n},
\end{align*}
where $L$ and $C$ are defined as in Theorem~\ref{thm:error_bound}.
\end{theorem}

\textbf{Remarks.}
Our privacy-aware protocol ensures that:
(i)~The buyer never accesses individual models or raw data from the sellers;  
(ii)~sellers do not receive any information about other sellers' data or models;  
(iii)~buyers cannot reconstruct individual models by differencing merged ones; and  
(iv)~dishonest participants can be detected through the protocol’s mask consistency check.
A high-level overview of DMVM framework is illustrated in Figure~\ref{fig:Overview} in Appendix~\ref{mainalg}. The complete procedure for DMVM is presented in Algorithm~\ref{alg:dmvm} in Appendix~\ref{mainalg}. In the next section, we evaluate the performance of this framework by simulating data marketplaces in a multi-task setting.
 
\section{Numerical Experiments}
\label{sec:experiments}



This section investigates whether DMVM can serve as a reliable and efficient proxy for dataset valuation in practical multi-task settings. Since this is, to our knowledge, the first study on multi-task dataset valuation, there exists no established benchmark or baseline for comparison. Consequently, we design two types of realistic and controlled simulation environments that allow us to test whether our method can correctly recover the ground-truth ranking and values of sellers with respect to their ground-truth Shapley value. Here, we discuss the realistic scenario and the results for the controlled setting are provided in Appendix~\ref{controlled}, where we construct a spectrum of sellers with varying degrees of relevance to the buyer's target tasks.

\subsection{Experimental Setup.}
We simulate a decentralized marketplace with $n$ sellers, where each seller owns a dataset $\mathcal{D}_i$. The buyer is interested in a set of target tasks $\mathcal{T}$.
For natural language processing (\textbf{NLP}) experiments, we consider three task families, each represented by three datasets:
\begin{itemize}[leftmargin=*,noitemsep,topsep=0pt]
    \item \textbf{Review classification:} SST-2 \cite{wang-etal-2018-glue}, IMDb \cite{maas-EtAl:2011:ACL-HLT2011}, and Yelp \cite{zhang2015character}.
    \item \textbf{Paraphrase identification:} MRPC \cite{wang-etal-2018-glue}, QQP \cite{wang-etal-2018-glue}, and PAWS \cite{zhang2019paws}.
    \item \textbf{Question answering:} BoolQ \cite{clark2019boolq}, PubMedQA \cite{jin2019pubmedqa}, and StrategyQA \cite{geva2021did}.
\end{itemize}

Each dataset is treated as a seller. Thus, the NLP marketplace contains nine sellers in total. The buyer's target tasks can be any subset of these tasks, and the validation utility is computed by averaging performance across the corresponding task-level validation sets. This design captures a realistic setting in which multiple sellers provide data for the same task, but from different domains or distributions. For instance, SST-2, IMDb, and Yelp all correspond to review or sentiment classification, yet differ in text length, source domain, and linguistic style. We use \texttt{google/flan-t5-small} \cite{flan} as the shared pretrained model.

For \textbf{vision} experiments, we construct a marketplace using four task families:
\begin{itemize}[leftmargin=*,noitemsep,topsep=0pt]
    \item \textbf{Digit classification:} MNIST \cite{lecun1998gradient}, SVHN \cite{37648svhn}, USPS \cite{hull2002database}, and EMNIST \cite{cohen2017emnist}.
    \item \textbf{Animal classification:} CIFAR-10 cat/dog \cite{krizhevsky2009learning} and Oxford-IIIT Pets cat/dog \cite{parkhi12a}.
    \item \textbf{Scene classification:} SUN397 \cite{5539970sun} and MIT Indoor67 \cite{quattoni2009recognizing}.
    \item \textbf{Land-use classification:} EuroSAT \cite{helber2019eurosat} and RESISC45 \cite{cheng2017remote}.
\end{itemize}

Again, each dataset corresponds to a seller. This yields ten vision sellers in total. The datasets within each task family share the same high-level prediction objective, but differ in visual domain, image statistics, and label distribution. For the shared pretrained model, we use \texttt{openai/clip-vit-base-patch32} \cite{pmlr-v139-radford21a}. More implementation details are provided in Appendix~\ref{impledetails}. 
\subsection{Baselines}
\textbf{Dataset Shapley as ground truth (GT).}
Dataset Shapley serves as the oracle reference. For each seller $i$, it computes the exact Shapley value using Eq.~\eqref{gt-equ}, where each coalition utility is obtained by training on the corresponding subset of sellers' datasets. Since this requires access to all sellers' raw datasets, we use it only for evaluation.

\textbf{MC-Dataset-Shapley~\cite{pmlr-v97-ghorbani19c}.}
MC-Dataset-Shapley provides MC approximation of Dataset Shapley by sampling $m$ coalitions or permutations instead of enumerating all possible coalitions.

\textbf{DU-Shapley~\cite{garrido2024shapley}.}
DU-Shapley is a dataset-level Shapley approximation based on discrete-uniform sampling. For seller $i$, it estimates the value as $ \psi_i
    =
    \frac{1}{n}
    \sum_{k=0}^{n-1}
    \left[
    u\!\left(\mathcal{D}^{(k)} \cup \mathcal{D}_i\right)
    -
    u\!\left(\mathcal{D}^{(k)}\right)
    \right],
$
where $\mathcal{D}^{(k)}$ is a {dataset that its data samples are} sampled uniformly without replacement from $
\mathcal{D}_{-i}=\bigcup_{j\in [n]\setminus\{i\}} \mathcal{D}_j,
$ with size $k\mu_{-i}$, and $\mu_{-i}=\frac{1}{n-1}|\mathcal{D}_{-i}|$. {({$|\mathcal{D}_{-i}|$ is the size of data samples in $\mathcal{D}_{-i}$})}. In our multi-task adaptation, $u(\cdot)$ is the average validation performance defined in Eq.~\eqref{gt-utility}.

\textbf{Our methods.}
We report both DMVM and MC-DMVM as our proposed methods. In MC-DMVM, the MC sampling variant of DMVM introduced in Theorem~\ref{thm:mc_error_bound}, seller values are estimated from $m$ sampled marginal contributions instead of all coalitions. Table~\ref{tab:methods_comparison} illustrates the computational time and privacy aspect of each method. 
\subsection{Evaluation Metrics}
We evaluate each method from two complementary perspectives. First, we use rank-correlation metrics used in the literature \citep{wang2025data,yang2024inflation,tian2022private,lu2024data} to measure whether a method correctly identifies the relative ordering of sellers. Second, we use mean squared error as an end-to-end value-estimation metric to measure whether the estimated numerical values are close to the ground-truth Dataset Shapley values. 

\textbf{Kendall’s $\tau$} measures the fraction of \emph{concordant} versus \emph{discordant} seller pairs. For two sellers $i$ and $j$ with predicted values $(v_i, v_j)$ and ground-truth values $(g_i, g_j)$, the pair is concordant if $
(v_i - v_j)(g_i - g_j) > 0.
$
Let $n_c$ and $n_d$ denote the number of concordant and discordant pairs among all $\binom{n}{2}$ pairs. Then, $\tau$ is defined as $\tau = \frac{n_c - n_d}{\binom{n}{2}}.$
The score ranges from $-1$ (perfect disagreement) to $1$ (perfect agreement).

\textbf{Spearman’s $\rho$} computes the Pearson correlation between the predicted and ground-truth rank vectors. Let $r_i$ and $s_i$ denote the predicted and ground-truth ranks of seller $i$, respectively. Let $\bar{r}$ and $\bar{s}$ denote the corresponding average of ranks across all sellers. Then, $\rho$ is defined as
$
\rho = \frac{\sum_{i=1}^{n}(r_i - \bar{r})(s_i - \bar{s})}
{\sqrt{\sum_{i=1}^{n}(r_i - \bar{r})^2}\sqrt{\sum_{i=1}^{n}(s_i - \bar{s})^2}}.
$
When no rank ties exist, this simplifies to $
\rho = 1 - \frac{6\sum_{i=1}^{n} d_i^2}{n(n^2 - 1)}$, where $d_i = r_i - s_i$. $\rho = 1$ indicates perfect alignment, $\rho = -1$ complete reversal, and $\rho = 0$ no correlation. While Kendall’s $\tau$ emphasizes pairwise ordering consistency, Spearman’s $\rho$ captures overall rank alignment; using both provides a comprehensive evaluation of ranking quality.

\textbf{End-to-end value-estimation error.}
While rank correlations evaluate whether a method orders sellers correctly, they do not measure whether the estimated seller values are numerically accurate. Therefore, we also report the mean squared error (MSE) between the estimated seller values and the ground-truth Dataset Shapley values as $    \mathrm{MSE}
    =
    \frac{1}{n}
    \sum_{i=1}^{n}
    (
    \phi_i-\hat{\phi}_i
    )^2,
$
where $\phi_i$ is the ground-truth Dataset Shapley value of seller $i$, and $\hat{\phi}_i$ is the value estimated by the method being evaluated. We refer to this as an end-to-end metric because it evaluates the final output of the valuation pipeline: the numerical value assigned to each seller.

\begin{table}
\centering
\scriptsize
\caption{Performance of valuation methods on vision and NLP tasks. Buyer's target task size $|\mathcal{T}|=3$ for NLP and $|\mathcal{T}|=4$ for vision.}
\label{tab:dmvm_vision_nlp_results}
\resizebox{\textwidth}{!}{
\begin{tabular}{c|l|ccc|ccc}
\toprule
\multirow{2}{*}{\textbf{$n$}} 
& \multirow{2}{*}{\textbf{Method}} 
& \multicolumn{3}{c|}{\textbf{Vision Tasks}} 
& \multicolumn{3}{c}{\textbf{NLP Tasks}} \\
\cmidrule(lr){3-5} \cmidrule(lr){6-8}
& & \textbf{$\tau$} & \textbf{$\rho$} & \textbf{MSE}
  & \textbf{$\tau$} & \textbf{$\rho$} & \textbf{MSE} \\
\midrule

\multirow{2}{*}{7}
& DU-Shapley 
& $0.8305 \pm 0.1247$ & $0.9174 \pm 0.0752$ & $0.000770 \pm 0.000490$
& $0.7235 \pm 0.2117$ & $0.8207 \pm 0.1953$ & $0.003324 \pm 0.002471$ \\
& DMVM 
& $0.8079 \pm 0.1356$ & $0.8988 \pm 0.0895$ & $0.000756 \pm 0.000368$
& $0.7784 \pm 0.1912$ & $0.8675 \pm 0.1589$ & $0.001967 \pm 0.001442$ \\

\midrule
\multirow{2}{*}{8}
& DU-Shapley 
& $0.8384 \pm 0.1079$ & $0.9250 \pm 0.0667$ & $0.000612 \pm 0.000391$
& $0.7132 \pm 0.1869$ & $0.8213 \pm 0.1629$ & $0.002661 \pm 0.001996$ \\
& DMVM 
& $0.7955 \pm 0.1193$ & $0.8963 \pm 0.0772$ & $0.000761 \pm 0.000344$
& $0.7578 \pm 0.1739$ & $0.8613 \pm 0.1351$ & $0.001815 \pm 0.001294$ \\

\midrule
\multirow{2}{*}{9}
& DU-Shapley 
& $0.8410 \pm 0.0993$ & $0.9288 \pm 0.0607$ & $0.000476 \pm 0.000315$
& $0.7030 \pm 0.1726$ & $0.8179 \pm 0.1475$ & $0.002122 \pm 0.001628$ \\
& DMVM 
& $0.7795 \pm 0.1098$ & $0.8893 \pm 0.0727$ & $0.000747 \pm 0.000322$
& $0.7487 \pm 0.1539$ & $0.8617 \pm 0.1200$ & $0.001668 \pm 0.001150$ \\

\bottomrule
\end{tabular}
}
\end{table}
\begin{table}
\centering
\scriptsize
\caption{Comparison of valuation methods for different numbers of buyer's target tasks. Number of sellers $n=9$ for NLP and $n=10$ for vision.}
\label{tab:dmvm_target_tasks_results}
\resizebox{\textwidth}{!}{
\begin{tabular}{c|l|ccc|ccc}
\toprule
\multirow{2}{*}{\textbf{$|\mathcal{T}|$}} 
& \multirow{2}{*}{\textbf{Method}} 
& \multicolumn{3}{c|}{\textbf{Vision Tasks}} 
& \multicolumn{3}{c}{\textbf{NLP Tasks}} \\
\cmidrule(lr){3-5} \cmidrule(lr){6-8}
& & \textbf{$\tau$} & \textbf{$\rho$} & \textbf{MSE}
  & \textbf{$\tau$} & \textbf{$\rho$} & \textbf{MSE} \\
\midrule

\multirow{2}{*}{2}
& DU-Shapley
& $0.8094 \pm 0.1028$ & $0.9100 \pm 0.0722$ & $0.000651 \pm 0.000571$
& $0.7469 \pm 0.1464$ & $0.8657 \pm 0.1137$ & $0.002924 \pm 0.002667$ \\
& DMVM
& $0.7738 \pm 0.1827$ & $0.8677 \pm 0.1729$ & $0.001442 \pm 0.000962$
& $0.8441 \pm 0.1141$ & $0.9315 \pm 0.0616$ & $0.002240 \pm 0.001838$ \\

\midrule
\multirow{2}{*}{3}
& DU-Shapley
& $0.8367 \pm 0.0876$ & $0.9270 \pm 0.0529$ & $0.000472 \pm 0.000363$
& $0.7030 \pm 0.1726$ & $0.8179 \pm 0.1475$ & $0.002122 \pm 0.001628$ \\
& DMVM
& $0.7837 \pm 0.1404$ & $0.8788 \pm 0.1193$ & $0.000969 \pm 0.000487$
& $0.7487 \pm 0.1539$ & $0.8617 \pm 0.1200$ & $0.001668 \pm 0.001150$ \\

\midrule
\multirow{2}{*}{4}
& DU-Shapley
& $0.8399 \pm 0.0938$ & $0.9287 \pm 0.0567$ & $0.000383 \pm 0.000259$
& $0.7385 \pm 0.1504$ & $0.8530 \pm 0.1196$ & $0.001721 \pm 0.001088$ \\
& DMVM
& $0.7697 \pm 0.1086$ & $0.8842 \pm 0.0750$ & $0.000732 \pm 0.000296$
& $0.6856 \pm 0.2082$ & $0.8004 \pm 0.1910$ & $0.001383 \pm 0.000815$ \\

\midrule
\multirow{2}{*}{5}
& DU-Shapley
& $0.8397 \pm 0.0929$ & $0.9312 \pm 0.0554$ & $0.000329 \pm 0.000195$
& $0.7650 \pm 0.1211$ & $0.8792 \pm 0.0796$ & $0.001480 \pm 0.000755$ \\
& DMVM
& $0.7556 \pm 0.0940$ & $0.8822 \pm 0.0579$ & $0.000590 \pm 0.000198$
& $0.6980 \pm 0.1656$ & $0.8267 \pm 0.1496$ & $0.001211 \pm 0.000607$ \\

\bottomrule
\end{tabular}
}
\end{table}

\begin{figure}[t]
    \centering

    \begin{minipage}{0.29\textwidth}
        \centering
        \includegraphics[width=\linewidth]{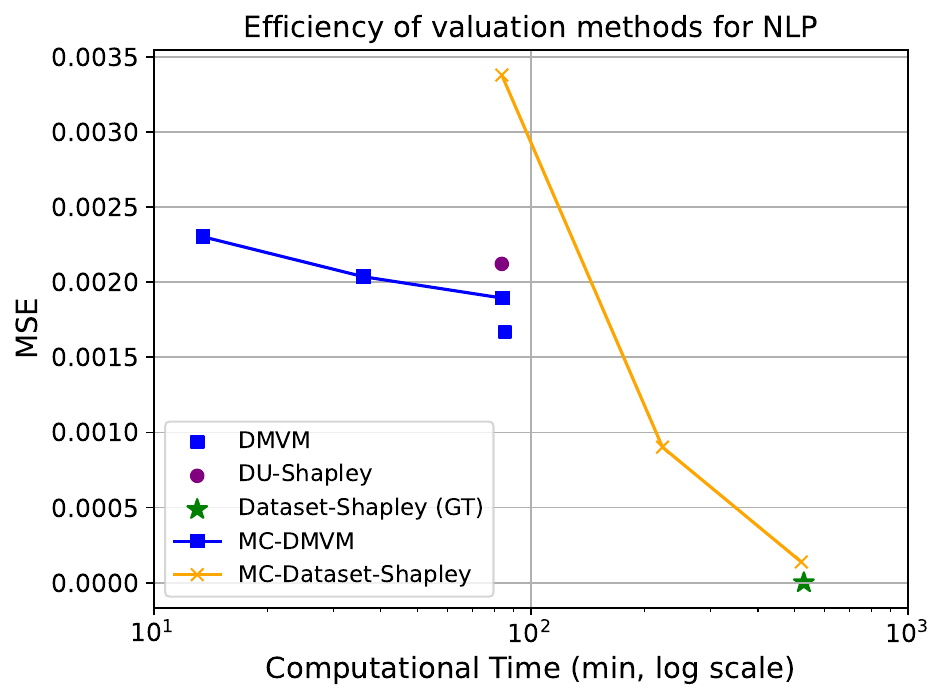}
    \end{minipage}
    \hspace{0.03\textwidth}
    \begin{minipage}{0.29\textwidth}
        \centering
        \includegraphics[width=\linewidth]{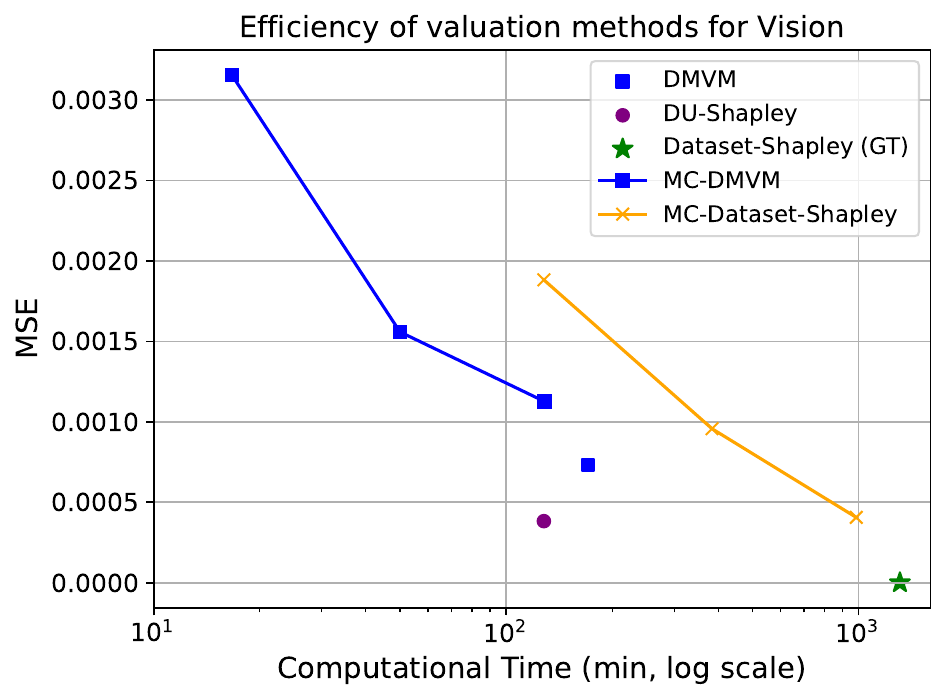}
    \end{minipage}

    \caption{Efficiency-accuracy trade-off of valuation methods on NLP and vision tasks. 
The $x$-axis reports wall-clock computational time in minutes, and the $y$-axis reports MSE relative to ground-truth.}
    \label{fig:two_figures}
\end{figure}

\subsection{Results and Discussion}

We evaluate whether DMVM can approximate ground-truth multi-task Dataset Shapley values while avoiding the two main limitations of existing valuation baselines: repeated retraining and access to sellers' private datasets. We emphasize that all non-DMVM baselines are privileged baselines: Dataset Shapley, MC-Dataset-Shapley, and DU-Shapley require access to sellers' datasets in order to train or evaluate models on seller coalitions. Therefore, they should be interpreted as practical upper-bound references rather than deployable baselines in decentralized data marketplaces. In contrast, DMVM estimates seller values only through merged models and does not require access to raw seller datasets.

Table~\ref{tab:dmvm_vision_nlp_results} evaluates DMVM as the number of sellers varies. Across both modalities, DMVM achieves strong agreement with the ground-truth Dataset Shapley values. In vision, DMVM obtains Kendall's $\tau$ values around $0.78$--$0.81$ and Spearman's $\rho$ values around $0.89$, with MSE on the order of $10^{-4}$. Although DU-Shapley gives slightly higher rank correlations in vision, DMVM remains close while operating under a substantially more realistic constraint. In NLP, DMVM consistently outperforms DU-Shapley in both ranking and MSE across all seller counts. For example, when $n=9$, DMVM improves Kendall's $\tau$ from $0.7030$ to $0.7487$, Spearman's $\rho$ from $0.8179$ to $0.8617$, and reduces MSE from $0.002122$ to $0.001668$.

Table~\ref{tab:dmvm_target_tasks_results} studies the effect of the number of buyer target tasks. DMVM remains stable across different values of $|\mathcal{T}|$ for both Vision and NLP experiments, suggesting that model merging provides a reliable proxy for multi-task valuation. In NLP, DMVM achieves lower MSE than DU-Shapley for all target-task sizes, indicating that it provides accurate end-to-end numerical value estimates even when the ranking metrics are sometimes slightly lower. All reported results are averaged over all possible choices of $n$ sellers and target task sets of size $|\mathcal{T}|$.

Figure~\ref{fig:two_figures} shows the efficiency-accuracy trade-off. Dataset Shapley gives the exact reference values but is computationally expensive because it requires training over all coalitions. MC-Dataset-Shapley reduces the number of coalitions but still requires retraining on sellers' datasets. In contrast, DMVM replaces coalition training with model merging and inference, making valuation substantially cheaper and compatible with decentralized settings. For the MC-based methods, the three plotted points correspond to $m\in\{9,24,56\}$ for NLP, and $m\in\{10,30,77\}$ for vision. Notably, the largest-sample MC-DMVM point matches the wall-clock time of DU-Shapley, while achieving lower MSE in NLP and competitive MSE in vision. This demonstrates that MC-DMVM provides a practical accuracy-efficiency trade-off: increasing $m$ reduces valuation error while keeping the method computationally lighter than retraining-based Shapley estimators.

More ablation studies on the convergence of MC-based methods as $m$ increases are provided in Appendix~\ref{convergence}.
Overall, these results show that DMVM provides accurate seller rankings and low end-to-end valuation error while avoiding the key limitations of existing baselines. Unlike Dataset Shapley, MC-Dataset-Shapley, and DU-Shapley, DMVM does not require access to sellers' datasets, making it both computationally efficient and privacy-aware for realistic multi-task data marketplaces.
Additional experiments with more model architectures and datasets for a systematic and controlled simulation, where the sellers might have datasets including more than one task in both overlapping and non-overlapping settings are provided in Appendix~\ref{controlled}. Scalability analysis is provided in Appendix~\ref{scalibility}. Moreover, we further investigate whether DMVM can identify and downweight harmful data sources by simulating sellers with increasing levels of label noise; the complete setup and analysis are presented in Appendix~\ref{harmf}.

\section{Conclusion}
This work reframes multi-task dataset valuation as a model-composition problem rather than a repeated retraining problem. In decentralized data marketplaces, the challenge is not only estimating which sellers are useful, but doing so before purchase, without assuming access to their private datasets. DMVM addresses this setting by using model merging as a proxy for coalition utility, enabling Shapley-style valuation through merged models instead of training on every seller subset. This preserves the economic interpretation of marginal contribution while making valuation more practical in multi-task and privacy-sensitive environments.
Our theoretical analysis characterizes the approximation gap between merging-based and ground-truth multi-task valuation, while our secure aggregation protocol prevents direct exposure of raw data or individual seller models. Empirically, DMVM and its MC variant achieve strong agreement with privileged retraining-based baselines across vision and NLP tasks, often attaining low value-estimation error with only inference-time computation. These results suggest that model merging can serve not only as a tool for multi-task adaptation, but also as an efficient mechanism for decentralized market valuation.

\bibliographystyle{plainnat}  
\bibliography{neurips_2026} 

\appendix

\section{Appendix}
\subsection{Algorithm}\label{mainalg}
\begin{algorithm}
\caption{DMVM: Decentralized Multi-task Valuation via Secure Model Merging}
\label{alg:dmvm}
\begin{algorithmic}[1]
\STATE \textbf{Input:} Buyer queries a potential group of sellers $[n]$, base model weights $\bm{W}_\text{pre}$, scaling hyperparameter $\alpha$
\FOR{each subset $\mathcal{S} \subseteq [n]$ with $|\mathcal{S}|\geq2$}
    \FOR{each layer pair $(\bm{W}^{(1)}, \bm{W}^{(2)})$}
        \IF{($\bm{W}^{(1)}, \bm{W}^{(2)}$) correspond to a 2-layer MLP}
            \STATE Sellers agree on a random permutation matrix $\bm{P}$.
            \STATE Define the layer transformation: $(\bm{\Psi}_1, \bm{\Psi}_2) = (\bm{P}, \bm{P}^\top)$.
        \ELSIF{($\bm{W}^{(1)}, \bm{W}^{(2)}$) are two consecutive linear layers}
            \STATE Sellers agree on a random scalar $\rho$.
            \STATE Define the layer transformation: $(\bm{\Psi}_1, \bm{\Psi}_2) = (\rho \bm{I}, \rho ^{-1}\bm{I})$.
        \ENDIF
        \FOR{each seller $i \in \mathcal{S}$}
            \IF{$\alpha=\frac{1}{|\mathcal{S|}}$}
            \STATE Generate masked shares:
            $\quad
            \bm{e}^{(i)}_1 =  \frac{\bm{\Psi}_1\bm{W}^{(1)}_i}{|\mathcal{S}|} - \bm{m}^{(1)}_i, \
            \bm{e}^{(i)}_2 = \frac{\bm{W}^{(2)}_i\bm{\Psi}_2}{|\mathcal{S}|}  - \bm{m}^{(2)}_i
            $
            \ELSE
            \STATE Compute local secrets:
            $\
            \bm{s}^{(1)}_i = \frac{\bm{W}^{(1)}_{\text{pre}}(1-|\mathcal{S}|\alpha)}{|\mathcal{S}|} + \alpha \bm{W}^{(1)}_i,\ \bm{s}^{(2)}_i = \frac{\bm{W}^{(2)}_{\text{pre}}(1-|\mathcal{S}|\alpha)}{|\mathcal{S}|} + \alpha \bm{W}^{(2)}_i
            $
            \STATE Generate masked shares:
            $\quad
            \bm{e}^{(i)}_1 = \bm{\Psi}_1 \bm{s}^{(1)}_i - \bm{m}^{(1)}_i, \
            \bm{e}^{(i)}_2 = \bm{s}^{(2)}_i \bm{\Psi}_2 - \bm{m}^{(2)}_i
            $
            \ENDIF
            \STATE Send $(\bm{e}^{(i)}_1, \bm{e}^{(i)}_2)$ to the buyer and $(\bm{m}^{(1)}_i, \bm{m}^{(2)}_i)$ to all other sellers.
        \ENDFOR
        \STATE Sellers compute aggregate masks: $\quad 
        \bm{m}^{(1)} = \sum_{i\in\mathcal{S}} \bm{m}^{(1)}_i, \quad \bm{m}^{(2)} = \sum_{i\in\mathcal{S}} \bm{m}^{(2)}_i
        $
        \STATE Each seller sends $(\bm{m}^{(1)}, \bm{m}^{(2)})$ to the buyer for verification.
        \STATE Buyer performs sanity check on consistency of $(\bm{m}^{(1)}, \bm{m}^{(2)})$.
        \STATE Buyer reconstructs the merged model weights:
        \begin{align*}\quad
        \bm{W}^{(1)}_{\text{merge}} = \bm{\bm{m}}^{(1)} + \sum_{i\in\mathcal{S}} \bm{e}^{(i)}_1, \ \bm{W}^{(2)}_{\text{merge}} = \bm{\bm{m}}^{(2)} + \sum_{i\in\mathcal{S}} \bm{e}^{(i)}_2
        \end{align*}
    \ENDFOR
\ENDFOR
\STATE \textbf{Output:} Privacy-aware merged models for all the subsets ready to use for valuation using Eq.~\eqref{eq:pri-shap} for each of the sellers.
\end{algorithmic}
\end{algorithm}
\clearpage

\begin{figure}[t]
    \centering

    \begin{minipage}{0.32\textwidth}
        \centering
        \includegraphics[width=\linewidth]{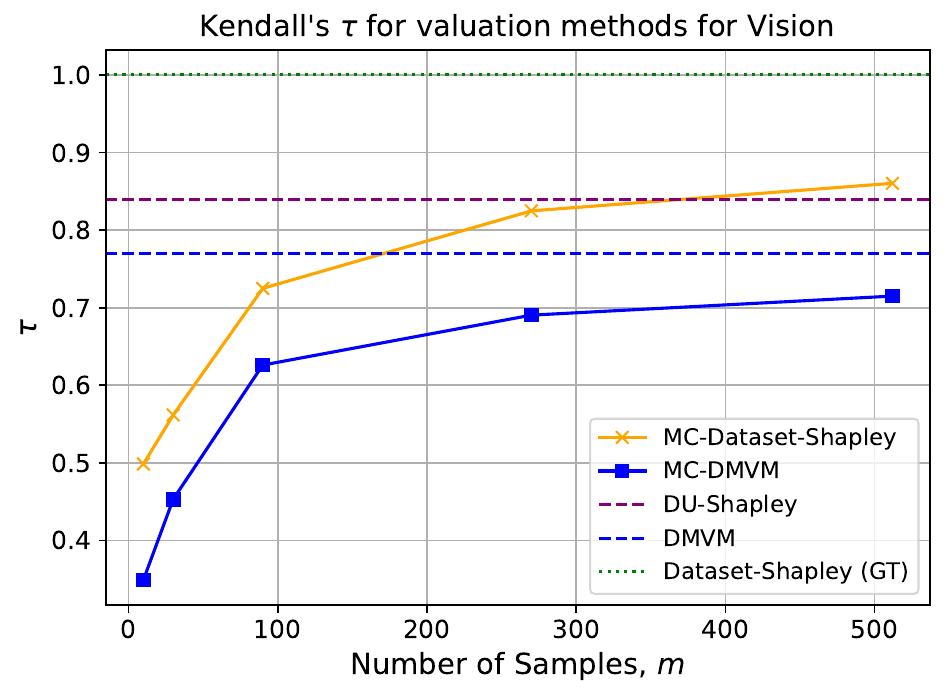}
    \end{minipage}
    \hfill
    \begin{minipage}{0.32\textwidth}
        \centering
        \includegraphics[width=\linewidth]{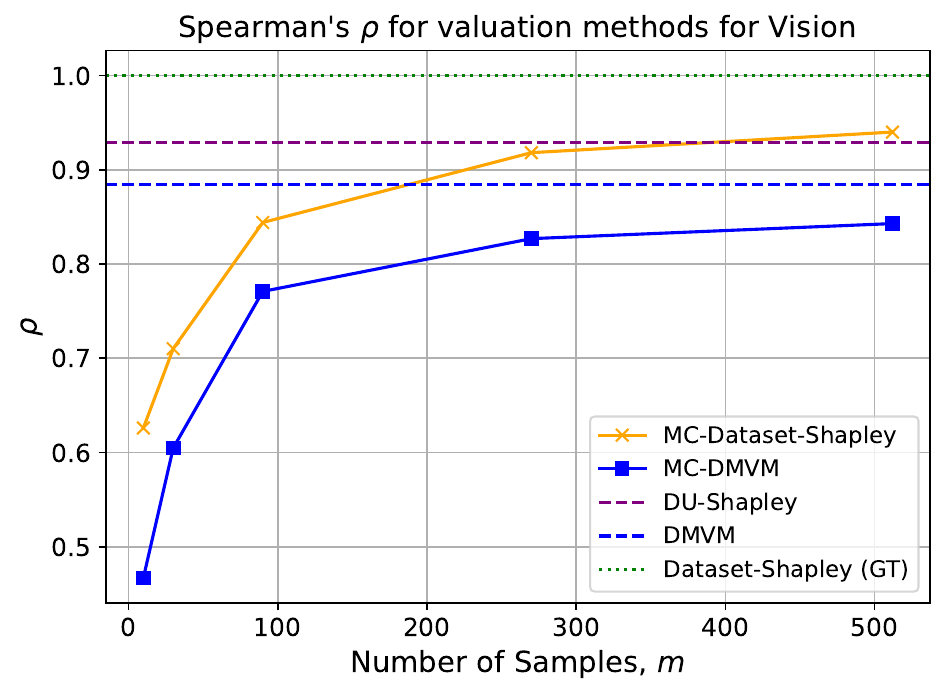}
    \end{minipage}
    \hfill
    \begin{minipage}{0.32\textwidth}
        \centering
        \includegraphics[width=\linewidth]{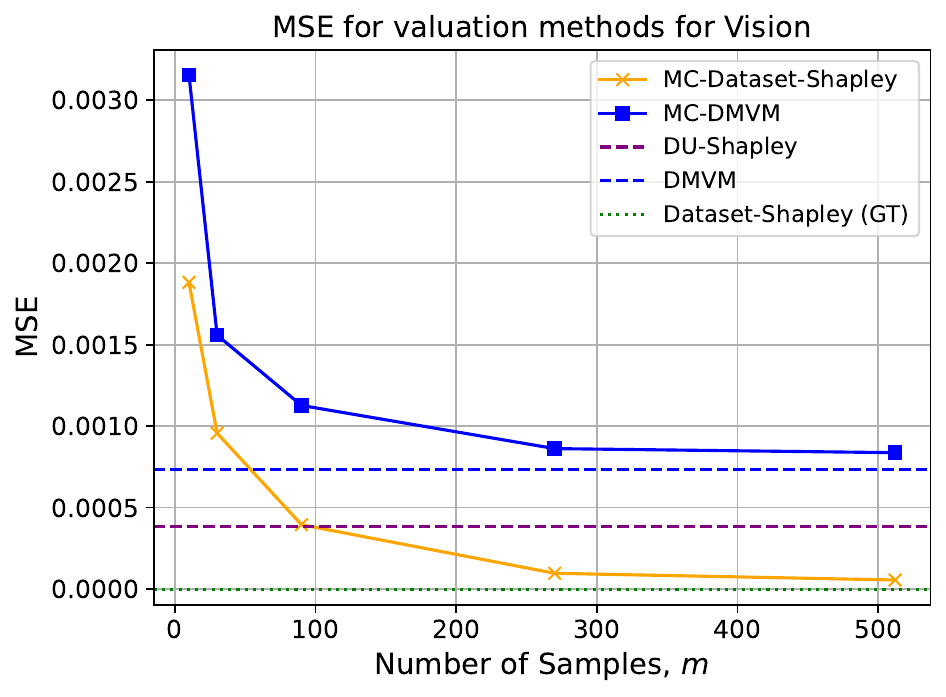}
    \end{minipage}

    \caption{Convergence behavior of MC-based valuation methods on vision tasks with $n=10$ sellers. The plots report Kendall's $\tau$, Spearman's $\rho$, and MSE as the number of MC samples $m$ increases. MC-Dataset-Shapley converges toward Dataset Shapley, while MC-DMVM converges toward deterministic DMVM.}
    \label{fig:convergence_vision}
\end{figure}

\begin{figure}[t]
    \centering

    \begin{minipage}{0.32\textwidth}
        \centering
        \includegraphics[width=\linewidth]{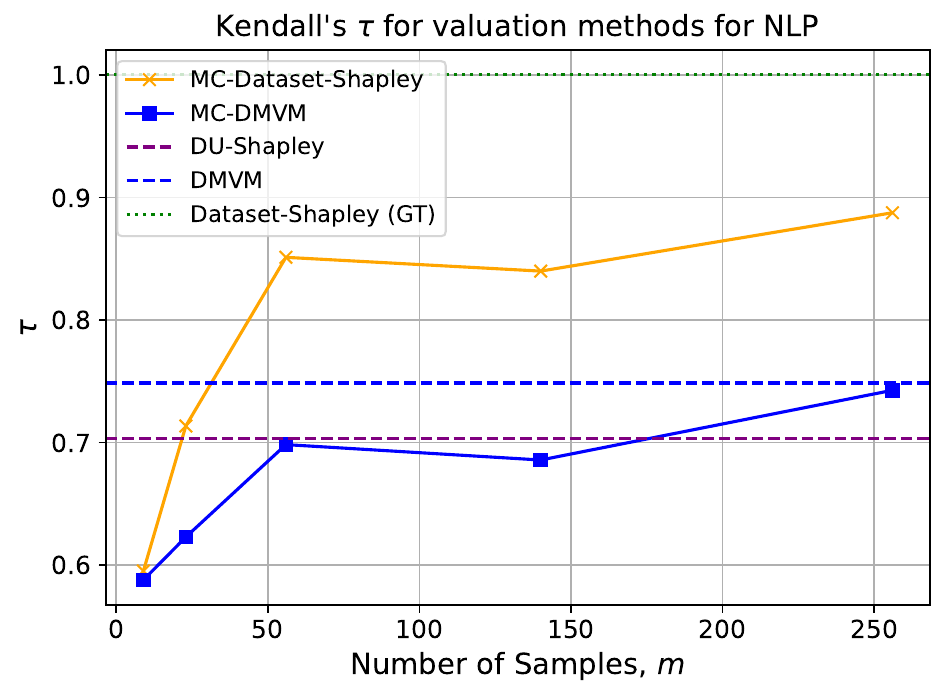}
    \end{minipage}
    \hfill
    \begin{minipage}{0.32\textwidth}
        \centering
        \includegraphics[width=\linewidth]{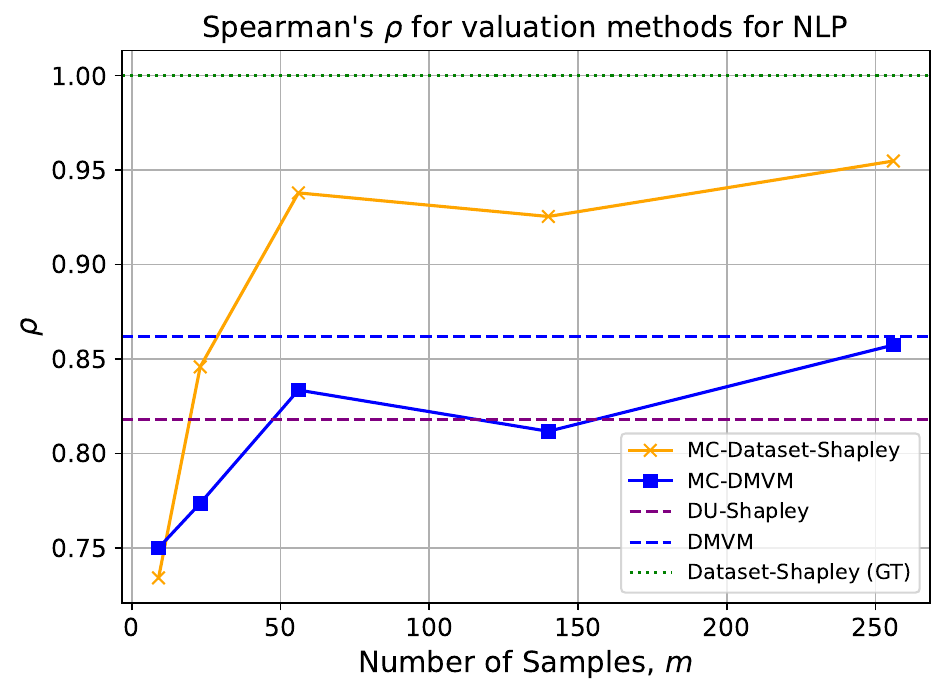}
    \end{minipage}
    \hfill
    \begin{minipage}{0.32\textwidth}
        \centering
        \includegraphics[width=\linewidth]{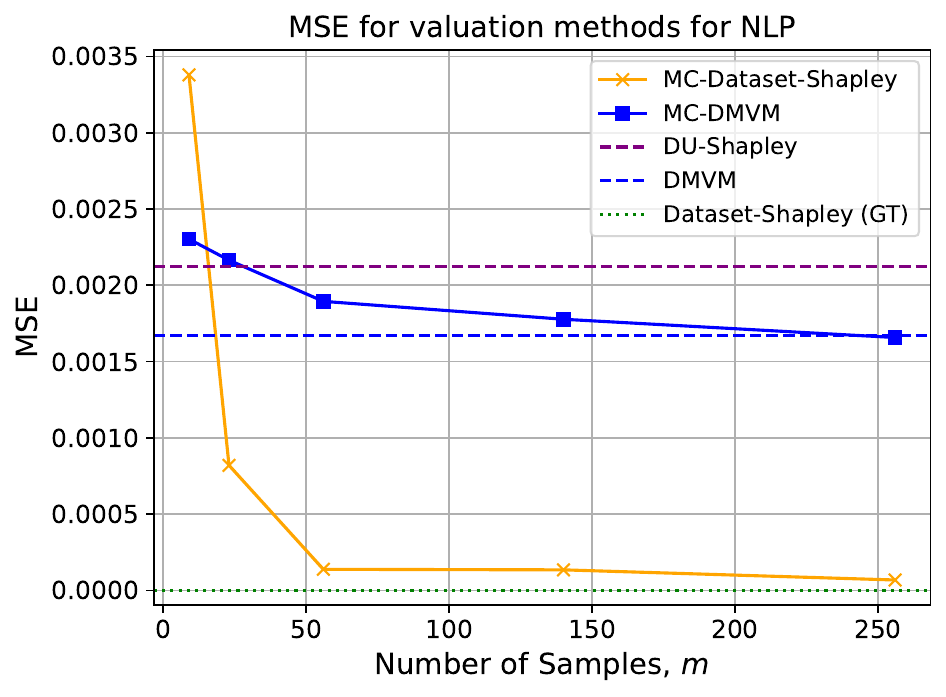}
    \end{minipage}
    \caption{Convergence behavior of MC-based valuation methods on NLP tasks with $n=9$ sellers. The plots report Kendall's $\tau$, Spearman's $\rho$, and MSE as the number of MC samples $m$ increases. MC-Dataset-Shapley approaches Dataset Shapley with larger $m$, while MC-DMVM approaches deterministic DMVM using only merging-based utility estimates.}
    \label{fig:convergence_nlp}
\end{figure}
\subsection{Convergence Analysis of MC-based methods}\label{convergence}

We further study the convergence behavior of the MC variants as the number of sampled coalitions $m$ increases. We focus on the setting with $n=10$ sellers for vision and $n=9$ sellers for NLP. For vision, we evaluate $m\in\{10,30,90,270,512\}$, and for NLP, we evaluate $m\in\{9,23,56,140,256\}$. The goal is to examine whether the MC estimators approach their corresponding deterministic counterparts as $m$ increases, while using only a fraction of the full coalition space.

Figures~\ref{fig:convergence_vision} and~\ref{fig:convergence_nlp} show that both MC-Dataset-Shapley and MC-DMVM improve consistently as $m$ increases. MC-Dataset-Shapley converges toward Dataset Shapley: its MSE decreases substantially, and both Kendall's $\tau$ and Spearman's $\rho$ improve with more samples. This is expected, since increasing $m$ gives a better MC approximation to the exact Shapley value. However, MC-Dataset-Shapley still requires access to sellers' datasets and repeated coalition training.

MC-DMVM converges toward deterministic DMVM, since it samples marginal contributions using the merging-based utility proxy rather than the retraining-based utility. In both vision and NLP, increasing $m$ reduces the gap between MC-DMVM and DMVM, especially in terms of MSE. The ranking metrics show small non-monotonic fluctuations, which are expected for sampling-based estimators, but they generally stabilize near the deterministic DMVM values as $m$ becomes larger.

These results support the error decomposition in Theorem~\ref{thm:mc_error_bound}: increasing $m$ reduces the statistical sampling error, while the remaining gap is governed by the deterministic merging approximation. Importantly, MC-DMVM provides two levels of computational savings. First, it replaces coalition training with model merging and inference. Second, it reduces the number of required coalition evaluations from exponential in the number of sellers, i.e., $2^{n-1}$, to only $m$ sampled evaluations. Therefore, MC-DMVM provides a practical accuracy-efficiency trade-off for decentralized valuation, achieving stable seller rankings and low value-estimation error without raw-data access or repeated retraining.

\subsection{Lemma}
\begin{lemma}[Convex Combination Upper Bound]\label{lemma}
Let $a_1, a_2, \dots, a_n \in \mathbb{R}$ and let $w_1, w_2, \dots, w_n \ge 0$ satisfy $
\sum_{i=1}^n w_i = b.
$
Then the weighted sum is bounded by the largest element, i.e.,
\begin{align*}
\sum_{i=1}^n w_i a_i \;\le\; b\max_{1 \le i \le n} a_i \;\le\; b\sup_{1 \le i \le n} a_i.
\end{align*}
\end{lemma}
\subsection{Error bound of our method}\label{proofthm1}

\citet{zhou2025task} proposed the following error bound between multi-task learning and task-arithmetic merging:
\begin{align*}
\|\bm{\theta}^{\text{merge}}_{\mathcal{S}}-\bm{\theta}^{\text{MTL}}_{\mathcal{S}}\|_2\leq s(\alpha s+1)C,
\end{align*}
where $s$ is the cardinality of $\mathcal{S}$, i.e., $s=|\mathcal{S}|$, and $C=\binom{h+2}{2} H_{\max} G_{\max}$ is a constant, where $h$ denotes the number of optimization steps (i.e., epochs), 
and $G_{\max}$ and $H_{\max}$ are upper bounds on the gradient and Hessian norms, respectively and $\alpha$ is the hyperparameter for task-arithmetic.
Assuming the loss function is L-Lipschitz continuous, we have:
\begin{align}\label{inquuu}
     |\mathcal{L}(\bm{\theta}^{\text{merge}}_{\mathcal{S}})-\mathcal{L}(\bm{\theta}^{\text{MTL}}_{\mathcal{S}})|\leq L\|\bm{\theta}^{\text{merge}}_{\mathcal{S}}-\bm{\theta}^{\text{MTL}}_{\mathcal{S}}\|_2\leq Ls(\alpha s+1)C.
\end{align}

Assuming that utility $u(\cdot)$ has a negative relation with loss function, we set $u=-\mathcal{L}$, then the error between our method and ground-truth Shapely will be:
\begin{align*}
    |\phi_i-\hat{\phi}_i|&=\frac{1}{n}\left|\sum_{\mathcal{S}\subseteq[n]\setminus\{i\}} \frac{u(\mathcal{S} \cup \{i\})-u(\mathcal{S)}-\hat{u}(\mathcal{S} \cup \{i\})+\hat{u}(\mathcal{S)}}{\binom{n-1}{|\mathcal{S}|}} \right| \\
    &\leq\frac{1}{n}\sum_{\mathcal{S}\subseteq[n]\setminus\{i\}} \frac{\left|u(\mathcal{S} \cup \{i\})-u(\mathcal{S)}-\hat{u}(\mathcal{S} \cup \{i\})+\hat{u}(\mathcal{S)}\right|}{\binom{n-1}{|\mathcal{S}|}}  \\
    &\leq\frac{1}{n}\sum_{\mathcal{S}\subseteq[n]\setminus\{i\}} \frac{\left|-\mathcal{L}({\bm{\theta}}_{\mathcal{S} \cup \{i\}}^{\text{MTL}})+\mathcal{L}({\bm{\theta}}_{\mathcal{S}}^{\text{MTL}})+{\mathcal{L}}({\bm{\theta}}_{{\mathcal{S} \cup \{i\}}}^{\text{merge}})-\mathcal{L}({\bm{\theta}}_{\mathcal{S}}^{\text{merge}})\right|}{\binom{n-1}{|\mathcal{S}|}}\\
    &\leq\frac{1}{n}\sum_{\mathcal{S}\subseteq[n]\setminus\{i\}} \frac{\left|{\mathcal{L}}({\bm{\theta}}_{{\mathcal{S} \cup \{i\}}}^{\text{merge}})-\mathcal{L}({\bm{\theta}}_{\mathcal{S} \cup \{i\}}^{\text{MTL}})\right|+\left|\mathcal{L}({\bm{\theta}}_{\mathcal{S}}^{\text{merge}})-\mathcal{L}({\bm{\theta}}_{\mathcal{S}}^{\text{MTL}})\right|}{\binom{n-1}{|\mathcal{S}|}}\\
    &\overset{(a)}{\leq} Ln(\alpha n+1)C + L(n-1)(\alpha (n-1)+1)C=(2\alpha n^2 - 2\alpha n+2n+\alpha-1)LC.
\end{align*}
Inequality (a) follows from Lemma~\ref{lemma}. 
Since \(\mathcal{S}\subseteq[n]\setminus\{i\}\), we have
\(|\mathcal{S} \cup \{i\}|\leq n\) and \(|\mathcal{S}|\leq n-1\). Hence, the two loss-difference terms are uniformly bounded by
\(Ln(\alpha n+1)C\) and \(L(n-1)(\alpha(n-1)+1)C\), respectively, using inequality~\eqref{inquuu}. Moreover, the Shapley coefficients form a convex combination, i.e., $
\frac{1}{n}
\sum_{\mathcal{S}\subseteq[n]\setminus\{i\}}
\frac{1}{\binom{n-1}{|\mathcal{S}|}}
=1.
$
Therefore, applying Lemma~\ref{lemma} with \(b=1\) yields inequality (a).

\subsection{Error bound of our privacy-aware method}\label{proofthm2}
Assuming that we want to exclude subsets of size less than $k$. Let $R$ be the maximum marginal utility, i.e., $\forall \mathcal{S}\subseteq [n]\setminus\{i\},\quad  |u(\mathcal{S} \cup \{i\})-u(\mathcal{S)}|\leq R$. Then, the error will be:
\begin{align*}
    |\phi_i-\tilde{\phi}_i|&=\frac{1}{n}\Bigg|\sum_{\mathcal{S}\subseteq[n]\setminus\{i\},|S|\geq k} \frac{u(\mathcal{S} \cup \{i\})-u(\mathcal{S)}-\hat{u}(\mathcal{S} \cup \{i\})+\hat{u}(\mathcal{S)}}{\binom{n-1}{|\mathcal{S}|}}\\&\quad+\sum_{\mathcal{S}\subseteq[n]\setminus\{i\},|S|< k} \frac{u(\mathcal{S} \cup \{i\})-u(\mathcal{S)}}{\binom{n-1}{|\mathcal{S}|}} \Bigg|\\
    &\leq\frac{1}{n}\sum_{\mathcal{S}\subseteq[n]\setminus\{i\},|S|\geq k} \Big|\frac{u(\mathcal{S} \cup \{i\})-u(\mathcal{S)}-\hat{u}(\mathcal{S} \cup \{i\})+\hat{u}(\mathcal{S)}}{\binom{n-1}{|\mathcal{S}|}}\Big|\\&\quad+\frac{1}{n}\sum_{\mathcal{S}\subseteq[n]\setminus\{i\},|S|< k} \Big|\frac{u(\mathcal{S} \cup \{i\})-u(\mathcal{S)}}{\binom{n-1}{|\mathcal{S}|}} \Big|\\
    & \leq\frac{1}{n}\sum_{\mathcal{S}\subseteq[n]\setminus\{i\},|S|\geq k} \frac{\left|{\mathcal{L}}({\bm{\theta}}_{{\mathcal{S} \cup \{i\}}}^{\text{merge}})-\mathcal{L}({\bm{\theta}}_{\mathcal{S} \cup \{i\}}^{\text{MTL}})\right|+\left|\mathcal{L}({\bm{\theta}}_{\mathcal{S}}^{\text{merge}})-\mathcal{L}({\bm{\theta}}_{\mathcal{S}}^{\text{MTL}})\right|}{\binom{n-1}{|\mathcal{S}|}}\\&\quad+\frac{1}{n}\sum_{\mathcal{S}\subseteq[n]\setminus\{i\},|S|< k}\frac{|\mathcal{L}({\bm{\theta}}_{\mathcal{S}}^{\text{MTL}})-\mathcal{L}({\bm{\theta}}_{\mathcal{S} \cup \{i\}}^{\text{MTL}})|}{\binom{n-1}{|\mathcal{S}|}}\\
    &\overset{(a)}{\leq} \frac{(n-k)}{n}(2\alpha n^2 - 2\alpha n+2n+\alpha-1)LC+\frac{k}{n}R.
\end{align*}
Inequality (a) follows by applying Lemma~\ref{lemma} separately to the two sums. For the first sum, the same argument as in Appendix~\ref{proofthm1} gives the uniform bound:
\[
\left|{\mathcal{L}}({\bm{\theta}}_{{\mathcal{S} \cup \{i\}}}^{\text{merge}})
-\mathcal{L}({\bm{\theta}}_{\mathcal{S} \cup \{i\}}^{\text{MTL}})\right|
+
\left|\mathcal{L}({\bm{\theta}}_{\mathcal{S}}^{\text{merge}})
-\mathcal{L}({\bm{\theta}}_{\mathcal{S}}^{\text{MTL}})\right|
\leq
\left(2\alpha n^2 - 2\alpha n+2n+\alpha-1\right)LC .
\]
The total weight of coalitions with \(|\mathcal{S}|\geq k\) is
$
\frac{1}{n}
\sum_{\substack{\mathcal{S}\subseteq[n]\setminus\{i\}\\ |\mathcal{S}|\geq k}}
\frac{1}{\binom{n-1}{|\mathcal{S}|}}
=
\frac{1}{n}
\sum_{s=k}^{n-1}1
=
\frac{n-k}{n}.
$
For the excluded coalitions with \(|\mathcal{S}|<k\), the marginal utility is bounded by \(R\), and their total weight is:
$
\frac{1}{n}
\sum_{\substack{\mathcal{S}\subseteq[n]\setminus\{i\}\\ |\mathcal{S}|< k}}
\frac{1}{\binom{n-1}{|\mathcal{S}|}}
=
\frac{1}{n}
\sum_{s=0}^{k-1}1
=
\frac{k}{n}.
$
Applying Lemma~\ref{lemma} with these two values of \(b\) gives inequality (a).
\subsection{MC Sampling approximation proof}\label{montecarl}
Let \(\phi_i\) denote the true Shapley value, \(\hat{\phi}_i\) the deterministic
approximation with error
\[
|\hat{\phi}_i - \phi_i| \le (2\alpha n^2 - 2\alpha n + 2n + \alpha - 1)\, L C,
\]
and \(\bar{\phi}_i^{(m)}\) the MC estimator of \(\hat{\phi}_i\) based on
\(m\) samples.  
Assume each sampled marginal contribution lies in an interval of width \(R\), i.e., $\forall \mathcal{S}\subseteq [n]\setminus\{i\},\quad |\hat{u}(\mathcal{S} \cup \{i\})-\hat{u}(\mathcal{S)}|\leq R$. 

By Hoeffding’s inequality, for any \(\varepsilon > 0\),
\[
\Pr\!\left( |\bar{\phi}_i^{(m)} - \hat{\phi}_i| \ge \varepsilon \right)
\le
2 \exp\!\left(
-\frac{2 m \varepsilon^2}{R^2}
\right).
\]

Using the triangle inequality,
\[
|\bar{\phi}_i^{(m)} - \phi_i|
\le
|\bar{\phi}_i^{(m)} - \hat{\phi}_i|
+
|\hat{\phi}_i - \phi_i|
\le
|\bar{\phi}_i^{(m)} - \hat{\phi}_i| + (2\alpha n^2 - 2\alpha n + 2n + \alpha - 1)\, L C.
\]

Thus, with probability at least \(1-\delta\), we have:
\[
|\bar{\phi}_i^{(m)} - \phi_i|
\le
(2\alpha n^2 - 2\alpha n + 2n + \alpha - 1)\, L C
+
\sqrt{
  \frac{R^2}{2m}\,
  \log\!\left(\frac{2}{\delta}\right)
}.
\]

\subsection{Controlled Simulations}\label{controlled}

The experiments in Section~\ref{sec:experiments} evaluate DMVM in practical marketplace scenarios where sellers correspond to different datasets and domains. However, in such settings, the true relevance of each seller to the buyer's target tasks is not directly controllable, since datasets may differ along several factors simultaneously, including domain, distribution, label space, and sample complexity. Therefore, we further conduct controlled simulations in which the relationship between each seller's data and the buyer's target tasks is explicitly designed. These experiments allow us to test whether DMVM can recover known relevance patterns, including highly useful, partially useful, irrelevant, and harmful sellers.

We consider two complementary experimental regimes.
First, we study \emph{single-domain multi-task} vision benchmarks, which provide an intuitive and interpretable setting for analyzing seller relevance.
Second, we extend our evaluation to \emph{multi-domain multi-task} experiments in both vision and NLP, where tasks may correspond to distinct domains. Detailed task distribution schemes for different buyer task set sizes and seller constructions are provided in Appendix~\ref{schemes}. Scalability analysis is provided in Appendix~\ref{scalibility}.

\subsubsection{Single-Domain Multi-Task Vision Experiments}
\label{sec:vision}
To establish intuition, we begin with vision experiments where all tasks belong to a single domain.
We consider MNIST and CIFAR-10, and define the buyer’s objective as learning a subset of classes, treated as multiple target tasks.
Multiple sellers provide datasets with varying relevance to the buyer’s task set.
For these experiments, we compute exact multi-task Shapley values by explicitly training models on all subsets of sellers.
The resulting Shapley values serve as the reference ordering against which DMVM-induced rankings are compared.

\paragraph{Non-Overlapping Sellers.}
We first consider a non-overlapping seller setting, where each seller provides data for a disjoint subset of classes.
As an illustrative MNIST example, we define the buyer’s target tasks as learning to recognize digits 0-6, and construct four sellers as follows:
\begin{itemize}
[leftmargin=*,noitemsep,topsep=0pt]
    \item Seller 1: classes (0, 1, 2, 3)
    \item Seller 2: classes (4, 5, 6)
    \item Seller 3: classes (7, 8, 9)
    \item Seller 4: mislabeled or noisy samples
\end{itemize}
Sellers 1 and 2 contribute directly to the buyer’s tasks, with Seller~1 expected to have higher Shapley value due to broader task coverage.
Seller~3 provides irrelevant data, while Seller~4 is harmful.
The reference ranking is therefore induced by the corresponding Shapley values.
\paragraph{Overlapping Sellers.}
We next consider an overlapping seller setting, where sellers provide partially intersecting subsets of the buyer’s tasks.
For MNIST, the buyer’s tasks correspond to recognizing digits 0-3, and we simulate five sellers with progressively decreasing overlap:
\begin{itemize}
[leftmargin=*,noitemsep,topsep=0pt]
    \item Seller 1: classes (0, 1, 2, 3)
    \item Seller 2: classes (0, 1, 2, 4)
    \item Seller 3: classes (0, 1, 4, 5)
    \item Seller 4: classes (0, 4, 5, 6)
    \item Seller 5: classes (4, 5, 6, 7)
\end{itemize}
This setting induces a smooth spectrum of Shapley values, allowing us to evaluate whether DMVM preserves relative ordering as task relevance decreases.

\subsubsection{Multi-Domain Multi-Task Language Modeling}
\label{sec:mdmt-nlp}
We next evaluate DMVM in a multi-domain multi-task language modeling setting, where each task corresponds to a distinct reasoning domain and models are generative.
We consider the following benchmarks:
\texttt{CommonsenseQA}~\citep{talmor-etal-2019-commonsenseqa} (commonsense multiple-choice reasoning),
\texttt{OpenBookQA}~\citep{OpenBookQA2018} (open-book science QA with multi-step reasoning),
\texttt{MedMCQA}~\citep{pmlr-v174-pal22a} (medical multiple-choice question answering),
\texttt{LogiQA}~\citep{liu2020logiqa} (logical reasoning over short passages),
\texttt{GSM8K}~\citep{cobbe2021gsm8k} (grade-school mathematical problem solving), and
\texttt{ARC-Easy}~\citep{allenai-arc} (elementary-level science question answering).
Each task is treated as a separate domain-specific learning objective.
We again adopt non-overlapping and overlapping seller designs. For non-overlapping,
given a buyer with $|\mathcal{T}|$ target tasks, sellers partition the task set disjointly.
For example, when the buyer has six target tasks, Seller~1 provides data for three tasks, Seller~2 for two tasks, and Seller~3 for the remaining task.
An additional seller provides data from an irrelevant task outside the buyer’s target set.
This design induces a clear hierarchy of Shapley values based on task coverage, enabling quantitative comparison with DMVM-induced rankings.
\paragraph{Results and Discussion.}
For single-domain multi-task vision experiments, we employed ViT-B/16 \citep{wu2020visual} as the shared pretrained model initialized from the HuggingFace checkpoint \texttt{google/vit-base-patch16-224-in21k}. Each seller fine-tunes this model for two epochs on its own dataset, after which we compute the privacy-aware valuation scores using $\tilde{\phi}$ in Eq.~\eqref{eq:pri-shap}. Fine-tuning is performed with AdamW using a learning rate of $1\times 10^{-4}$ and weight decay $0.01$, providing stable adaptation across all seller datasets. Experiments for multi-domain multi-task vision are provided in Appendix~\ref{multivismodel}. For multi-domain multi-task language experiments, we used two pretrained large language models with different scales and architectures:
\textbf{LLaMA-3.2-1B} \citep{grattafiori2024llama} and \textbf{Qwen2.5-3B} \citep{qwen25}. 
Tables~\ref{tab:nonoverlap}, \ref{tab:overlap}, and \ref{tab:nlp-res} collectively demonstrate that DMVM induces seller rankings closely aligned with ground-truth Shapley-value-based orderings across both vision and language modalities.
Performance is strongest in non-overlapping settings, while remaining robust under partial overlap and across diverse language domains.

In vision experiments, DMVM exactly recovers the Shapley ordering in non-overlapping settings (Table~\ref{tab:nonoverlap}), and remains highly correlated under partial overlap (Table~\ref{tab:overlap}).
In language modeling, DMVM achieves strong rank correlation for both LLaMA-3.2-1B and Qwen2.5-3B as the shared model, demonstrating that merging-based valuation extends naturally to generative, multi-domain scenarios (Table~\ref{tab:nlp-res}). Table~\ref{tab:wallclock-shapley} compares the wall-clock time required to compute dataset Shapley values using DMVM and exact ground-truth Shapley in the language domain, highlighting the computational efficiency of DMVM. 
\begin{table}[t]
\centering
\scriptsize
\caption{
Rank correlation between DMVM-induced rankings and multi-task Shapley-value-based rankings for single-domain multi-task vision experiments with non-overlapping sellers.
}
\label{tab:nonoverlap}
\begin{tabular}{l|cc|cc}
\toprule
\multirow{2}{*}{\(|\mathcal{T}|\)} & \multicolumn{2}{c|}{MNIST} & \multicolumn{2}{c}{CIFAR-10} \\ 
\cmidrule(lr){2-3} \cmidrule(lr){4-5}
 & \(\tau\) & \(\rho\) & \(\tau\) & \(\rho\) \\
\midrule
5 & 1.0 & 1.0 & 1.0  & 1.0  \\
6 & 1.0  & 1.0  & 1.0  & 1.0 \\
7 & 1.0  & 1.0  & 1.0  & 1.0  \\
\bottomrule
\end{tabular}
\end{table}
\begin{table}[t]
\centering
\scriptsize
\caption{
Rank correlation between DMVM-induced rankings and multi-task Shapley-value-based rankings for single-domain multi-task vision experiments with overlapping sellers.
}
\label{tab:overlap}
\begin{tabular}{l|cc|cc}
\toprule
\multirow{2}{*}{\(|\mathcal{T}|\)} & \multicolumn{2}{c|}{MNIST} & \multicolumn{2}{c}{CIFAR-10} \\ 
\cmidrule(lr){2-3} \cmidrule(lr){4-5}
 & \(\tau\) & \(\rho\) & \(\tau\) & \(\rho\) \\
\midrule
3 & 0.92 $\pm$ 0.16  & 0.95 $\pm$ 0.1 & 0.67 $\pm$ 0.0 & 0.8 $\pm$ 0.0 \\
4 & 0.8 $\pm$ 0.0 & 0.9 $\pm$ 0.0  & 0.85 $\pm$ 0.1 & 0.92 $\pm$ 0.05 \\
5 & 0.77 $\pm$ 0.13 & 0.87 $\pm$ 0.08  & 0.83 $\pm$ 0.2 & 0.87 $\pm$ 0.16\\
6 & 0.95 $\pm$ 0.1 & 0.97 $\pm$ 0.05 & 0.8 $\pm$ 0.16 & 0.88 $\pm$ 0.13 \\
\bottomrule
\end{tabular}
\end{table}
\begin{table}[t]
\centering
\scriptsize
\caption{
Rank correlation between DMVM-induced rankings and Shapley-value-based rankings for multi-domain multi-task language modeling.
}
\label{tab:nlp-res}
\begin{tabular}{l|cc|cc}
\toprule
\multirow{2}{*}{\(|\mathcal{T}|\)} & \multicolumn{2}{c|}{LLaMa-3.2-1B} & \multicolumn{2}{c}{Qwen2.5-3B} \\ 
\cmidrule(lr){2-3} \cmidrule(lr){4-5}
 & \(\tau\) & \(\rho\) & \(\tau\) & \(\rho\) \\
\midrule
3 (overlapping) & 0.78 & 0.87 & 0.84 & 0.9 \\
\midrule
5 & 0.89 & 0.93 & 1.0 & 1.0 \\
6 & 1.0 & 1.0  & 1.0  & 1.0 \\
\bottomrule
\end{tabular}
\end{table}
Taken together, these results demonstrate that DMVM remains effective across a wide range of shared model architectures, model scales, data modalities, and marketplace seller topologies, including both vision and language models. 

The consistent recovery of seller rankings indicates that DMVM is not tied to a specific architecture or parameter regime, but instead leverages transferable representation shifts induced by seller-specific fine-tuning. Importantly, DMVM achieves this while requiring only a single fine-tuning per seller, avoiding the combinatorial retraining costs associated with Shapley-value estimation and enabling scalable deployment in multi-seller settings. Moreover, valuation is performed through model merging rather than direct access to seller data, preserving data locality and supporting privacy-aware evaluation. 

\subsubsection{Multi-Domain Multi-Task Vision}
\label{multivismodel}

\begin{table}[h]
\centering
\small
\caption{
DMVM ranking performance in the multi-domain multi-task vision setting for different numbers of target tasks under non-overlapping and overlapping task distributions.
}
\label{tab:multivision}
\begin{tabular}{l|cc|cc}
\toprule
\multirow{2}{*}{\(|\mathcal{T}|\)} & \multicolumn{2}{c|}{Non-overlapping} & \multicolumn{2}{c}{Overlapping} \\ 
\cmidrule(lr){2-3} \cmidrule(lr){4-5}
 & \(\tau\) & \(\rho\) & \(\tau\) & \(\rho\) \\
\midrule
4 & 1.0 & 1.0 &  1.0  & 1.0  \\
5 & 1.0  &  1.0 & 1.0  & 1.0 \\
6 &  1.0 &  1.0 & 1.0  & 1.0  \\
\bottomrule
\end{tabular}
\end{table}

We evaluate DMVM in a multi-domain multi-task vision setting using \textbf{CLIP-ViT-B/32}~\citep{pmlr-v139-radford21a} as the shared pretrained model. To construct a set of heterogeneous vision tasks, we follow the dataset selection used in the Fusion benchmark~\citep{tang2024fusionbench}, which was designed for evaluating model merging across diverse visual domains.

Specifically, we consider:
MNIST~\citep{lecun1998gradient} (handwritten digit classification),
CIFAR-10~\citep{krizhevsky2009learning} (natural object classification),
GTSRB~\citep{6033395gts} (traffic sign recognition),
SVHN~\citep{37648svhn} (street-view house number digit classification),
Stanford Cars~\citep{6755945ca} (fine-grained car model classification),
EuroSAT~\citep{helber2019eurosat} (satellite land-use and land-cover classification), Oxford-IIIT~\citep{parkhi12a} (pet category), DTD~\citep{cimpoi2014describing} (describable texture classification), SUN397~\citep{5539970sun} (scene recognition across diverse environments), RESISC45~\citep{cheng2017remote} (Remote sensing image scene classification), Fashion-MNIST~\citep{xiao2017fashion} (clothing category), and Food101~\citep{bossard2014food} (food categories).
These datasets span multiple visual domains and data distributions, forming a heterogeneous multi-task setting suitable for evaluating multi-domain multi-task dataset valuation.

\paragraph{Results And Discussion.} As summarized in Table \ref{tab:multivision}, DMVM consistently recovers the ground-truth seller ranking with perfect agreement across all multi-domain settings we consider. Both correlations remain equal to 1.0,
under both non-overlapping and overlapping task-distribution schemes. This indicates that, even in the presence of substantial domain heterogeneity—ranging from digits and traffic signs to satellite imagery, textures, scenes, and fine-grained object categories—DMVM preserves the relative importance ordering of sellers exactly. The results demonstrate that our valuation proxy remains stable as the number of target tasks grows and as inter-seller task overlap increases, highlighting its robustness in realistic multi-domain multi-task vision scenarios.

\begin{table}[H]
\centering
\caption{Total wall-clock time required to compute dataset Shapley values using MTL (ground-truth) versus DMVM on language tasks. 
Measured averages: training = 1 hour, evaluation = 0.1 hour (6 minutes), merging (negligible $\approx$ 0).}
\label{tab:wallclock-shapley}
\begin{tabular}{c|cc}
\toprule
\# Sellers ($n$) & Dataset-Shapley (GT) Time (hours) & DMVM Time (hours) \\
\midrule
4 & $16 \times 1.1 = 17.6$ & $16\times 0.1 = 1.6$ \\
5 & $32 \times 1.1 = 35.2$ & $32 \times 0.1 = 3.2$ \\
\bottomrule
\end{tabular}
\end{table}

\subsection{Harmful Dataset Detection}\label{harmf}
\begin{figure}[h]
    \centering
    \includegraphics[width=0.4\linewidth]{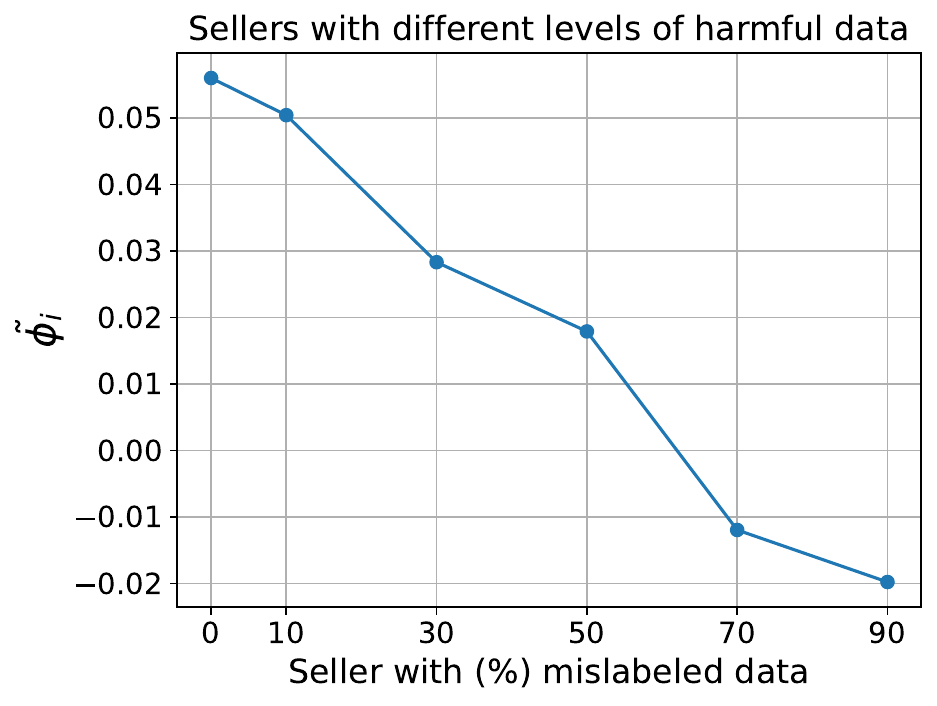}
    \caption{DMVM-induced shapley values of sellers with different percentages of mislabeled data. As label noise increases, the seller’s contribution decreases and eventually becomes negative, showing that DMVM successfully identifies and penalizes harmful data providers.}
    \label{fig:harmw}
\end{figure}
To examine whether DMVM can distinguish between helpful and harmful data providers, we design a controlled synthetic experiment in which a seller’s dataset is progressively degraded by randomly flipping an increasing fraction of its labels. This setup allows us to systematically vary data quality while keeping all other factors fixed. Figure~\ref{fig:harmw} illustrates the resulting Shapley value of the seller as a function of the mislabeling rate. We observe a clear and consistent monotonic decline in valuation as the noise level increases, indicating that DMVM is sensitive to degradation in data quality. This behavior highlights an important practical property of DMVM: beyond ranking relevant contributors highly, it can also reliably detect and penalize harmful or low-quality data sources in a multi-task learning setting.


\begin{figure}[h]
    \centering
    \includegraphics[width=0.63\linewidth]{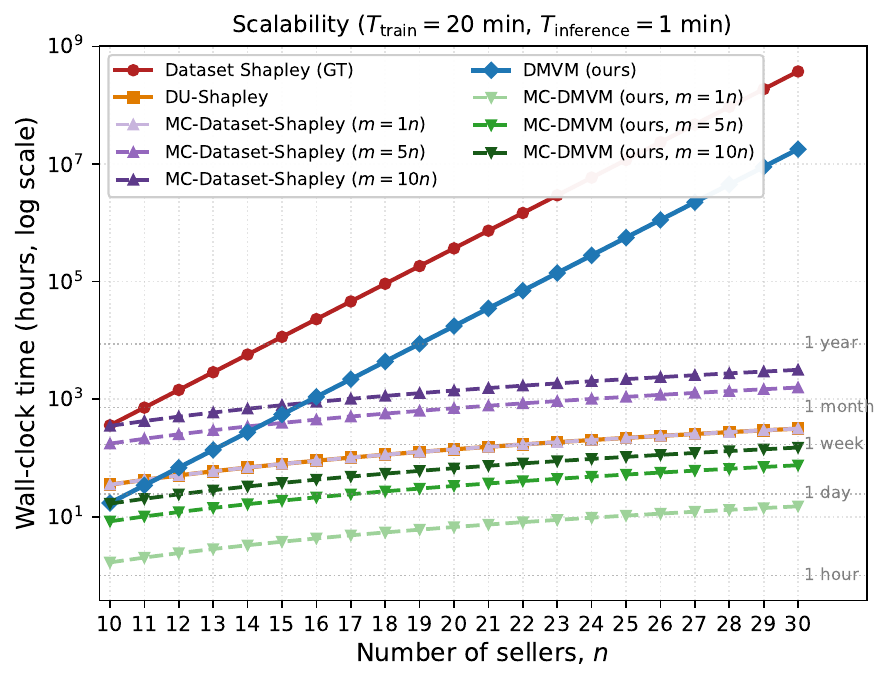}
    \caption{Projected wall-clock cost of each valuation method as the
marketplace size $n$ grows, computed from the asymptotic formulas in
Table~\ref{tab:methods_comparison} with $T_{\mathrm{train}}=20$ min and
$T_{\mathrm{inference}}=1$ min.}
    \label{fig:timescala}
\end{figure}
\subsection{Scalability Analysis}\label{scalibility}

To assess how DMVM scales beyond the regimes we measured directly, we
project the wall-clock cost of each method in
Table~\ref{tab:methods_comparison} using
$T_{\mathrm{train}}=20$ minutes and $T_{\mathrm{inference}}=1$ minute, a
reasonable assumption
for regular downstream tasks and $0.5$--$1$B parameter models. Figure~\ref{fig:timescala} reports the projected
runtime as the marketplace size grows from $n=10$ to $n=30$ sellers. For the
MC variants, we sweep $m\in\{n, 5n, 10n\}$ to span different possible sample budgets.
The figure shows two qualitatively different regimes. Exact Dataset Shapley
and deterministic DMVM both scale as $\mathcal{O}(2^n)$ and become
infeasible past $n\!\approx\!11$: at $n=30$, exact Shapley requires more
than $10^{8}$ hours of computation, while deterministic DMVM,
despite avoiding all training, still exceeds $10^{7}$ hours. The
polynomial methods are tractable in this regime, but DU-Shapley and
MC-Dataset-Shapley both require retraining on sellers' raw data, which is
incompatible with the decentralized setting we target. MC-DMVM is the only
method that is simultaneously polynomial in $n$, dependent only on
$T_{\mathrm{inference}}$, and privacy-aware: at $n=30$ it completes in
approximately $15$, $75$, and $150$ hours for $m=n$, $5n$, and $10n$,
respectively. Real-world data marketplaces could involve tens of
sellers per valuation round, and Figure~\ref{fig:timescala} shows that this is
precisely the regime where MC-DMVM remains practical while requiring a
single local fine-tuning per seller and no raw-data exchange.

\begin{figure}[h]
    \centering

    \begin{minipage}{0.32\textwidth}
        \centering
        \includegraphics[width=\linewidth]{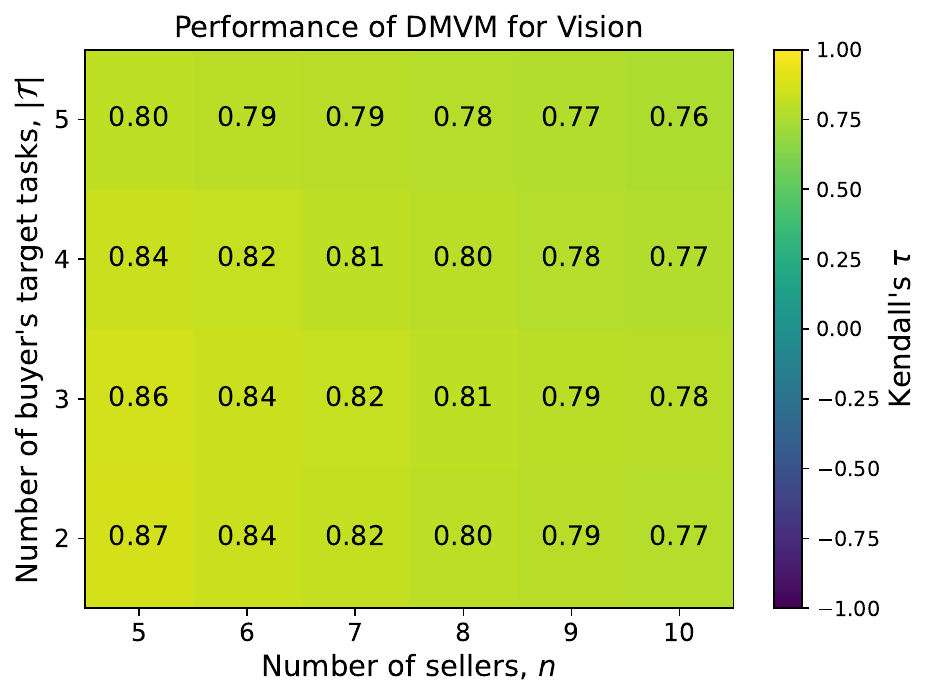}
    \end{minipage}
    \hfill
    \begin{minipage}{0.32\textwidth}
        \centering
        \includegraphics[width=\linewidth]{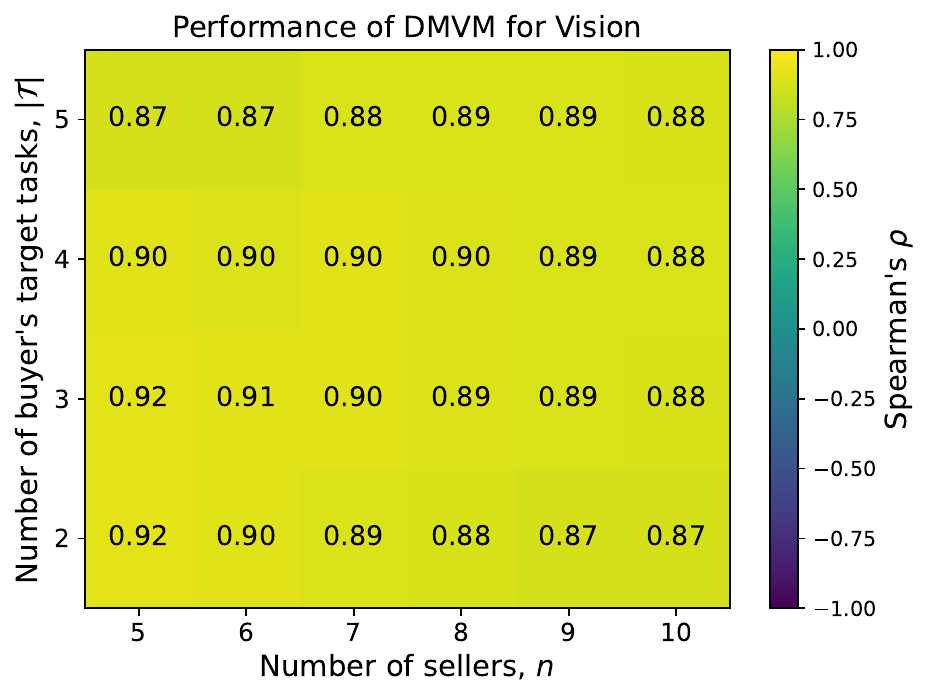}
    \end{minipage}
    \hfill
    \begin{minipage}{0.32\textwidth}
        \centering
        \includegraphics[width=\linewidth]{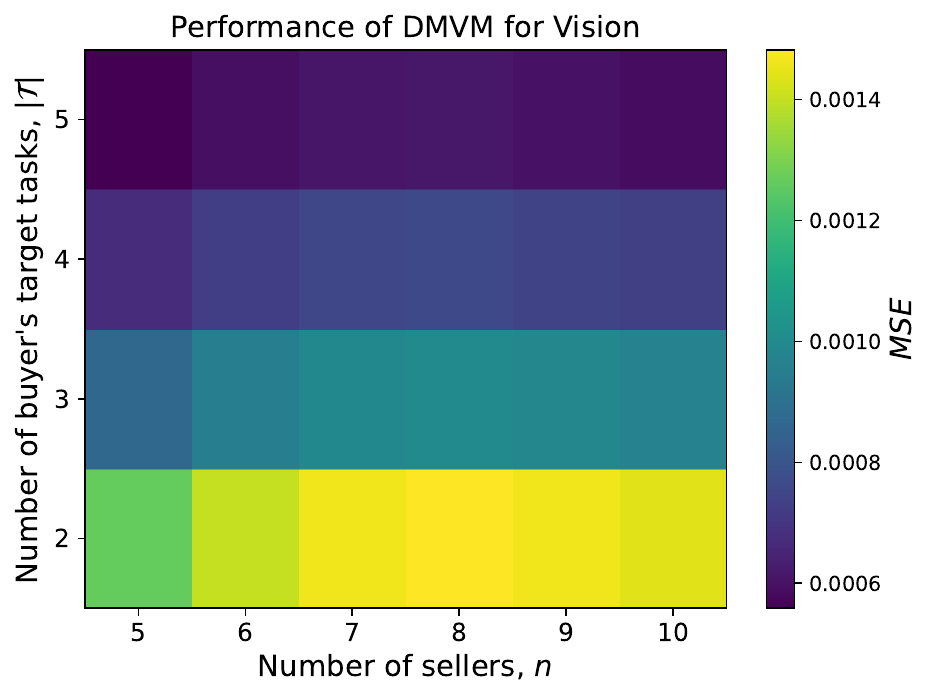}
    \end{minipage}

    \caption{DMVM performance on vision tasks across the full
$(n, |\mathcal{T}|)$ grid. Left to right: Kendall's $\tau$, Spearman's
$\rho$, and MSE relative to ground-truth Dataset Shapley.}
    \label{fig:vision_heatmap}
\end{figure}

\begin{figure}[h]
    \centering

    \begin{minipage}{0.32\textwidth}
        \centering
        \includegraphics[width=\linewidth]{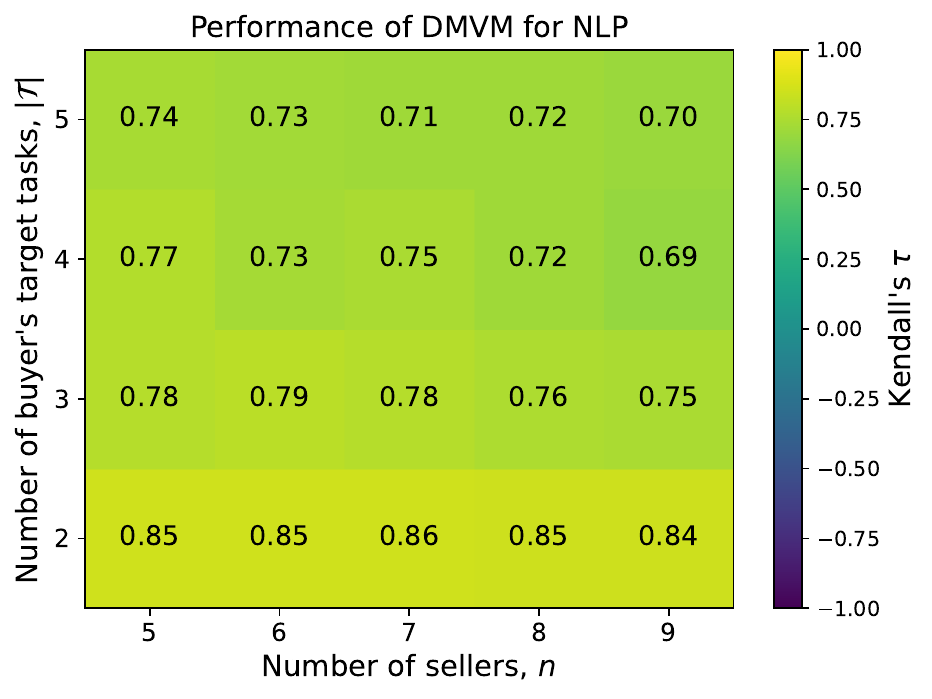}
    \end{minipage}
    \hfill
    \begin{minipage}{0.32\textwidth}
        \centering
        \includegraphics[width=\linewidth]{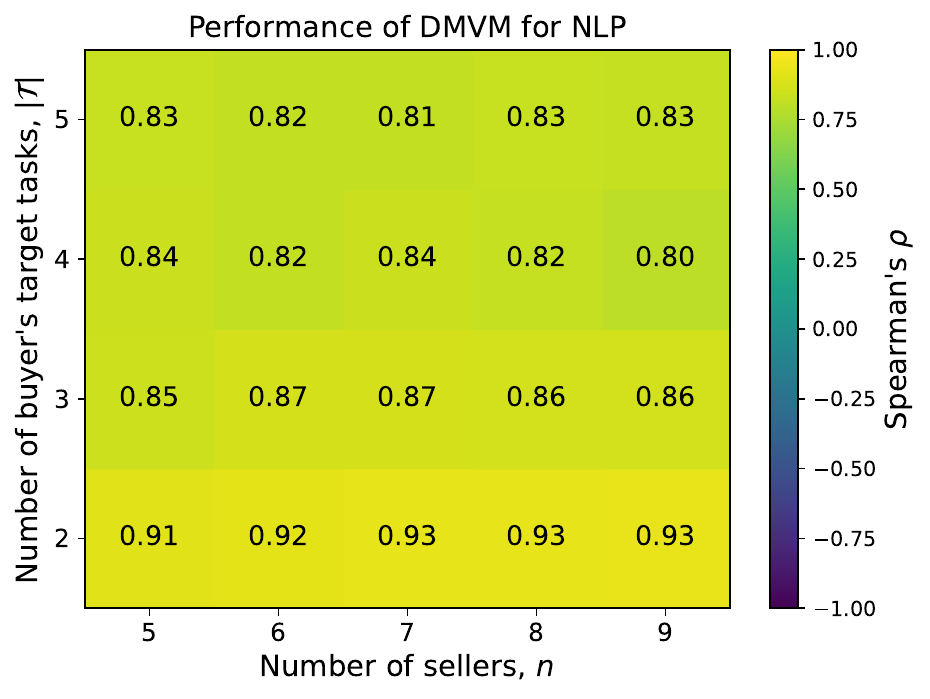}
    \end{minipage}
    \hfill
    \begin{minipage}{0.32\textwidth}
        \centering
        \includegraphics[width=\linewidth]{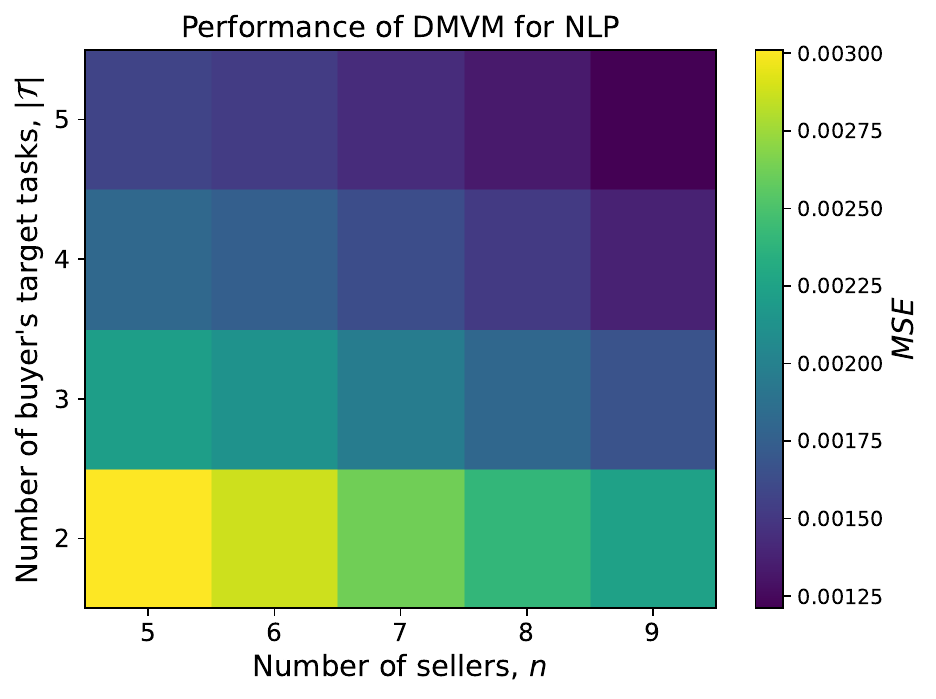}
    \end{minipage}

    \caption{DMVM performance on NLP tasks across the full
$(n, |\mathcal{T}|)$ grid. Left to right: Kendall's $\tau$, Spearman's
$\rho$, and MSE relative to ground-truth Dataset Shapley.}
    \label{fig:nlp_heatmap}
\end{figure}

Moreover, to better study the effect of scaling number of sellers, $n$, and number of buyer's target tasks, $|\mathcal{T}|$, on the valuation performance, we have provided an extended version of Table~\ref{tab:dmvm_vision_nlp_results} and Table~\ref{tab:dmvm_target_tasks_results}. Figures~\ref{fig:vision_heatmap}
and~\ref{fig:nlp_heatmap} report DMVM's empirical performance across the
full $(n, |\mathcal{T}|)$ grid, for vision and NLP, respectively.
Across all configurations, Kendall's $\tau$ remains above $0.69$, Spearman's
$\rho$ above $0.80$, and MSE stays on the order of $10^{-3}$ in both
modalities. MSE decreases as $|\mathcal{T}|$ grows, since averaging utility
over more validation tasks reduces variance in the estimated coalition values,
while rank-correlation metrics decline mildly with both $n$ and
$|\mathcal{T}|$: an expected effect, since larger marketplaces introduce more pairwise orderings to recover. The decline is
gradual rather than abrupt, indicating that DMVM remains stable across
the parameter range relevant to multi-task data marketplaces.

\subsection{Implementation Details}\label{impledetails}

All experiments are implemented in PyTorch. For both vision and NLP experiments, seller models are fine-tuned from a shared pretrained initialization using the Adam optimizer. Since computing the ground-truth Dataset Shapley values requires training models for an exponential number of seller coalitions, we use a lightweight fine-tuning setup and train most coalition models for only $2$--$3$ epochs. This makes the ground-truth computation feasible while still providing a consistent reference for comparing valuation methods.

For the NLP experiments, we use \texttt{google/flan-t5-small} as the shared pretrained model. The main NLP results are obtained using a batch size of $32$ and learning rate $3\times 10^{-4}$. For the vision experiments, we use \texttt{openai/clip-vit-base-patch32} as the shared pretrained model. The main vision results are obtained using a batch size of $64$, learning rate $10^{-5}$, and weight decay $10^{-4}$. For merging the seller models in each subset $\mathcal{S}$, we use task arithmetic with scaling coefficient $\alpha=\frac{1}{|\mathcal{S}|}$, corresponding to averaging the task vectors within the subset. For each buyer's target task, we evaluate utility on at most $2000$ validation examples, using the full test set when it contains fewer than $2000$ examples. The buyer's utility is then computed by averaging validation performance across the target tasks.
All main experiments can be run on a single NVIDIA H100 GPU. The main computational bottleneck is the retraining-based ground-truth and baseline computation, whereas DMVM only requires model merging over sellers and inference over the buyer's validation tasks.

\subsection{Task Distribution Schemes}\label{schemes}

In the controlled simulation experiments, for buyers with target task sets of varying cardinalities, we design task allocation schemes for both non-overlapping and overlapping seller constructions.
In the non-overlapping schemes, each fully relevant seller is assigned a distinct number of tasks from $\mathcal{T}$. This avoids ties or near-ties in Shapley rankings and ensures a clear separation in seller relevance. Sellers with irrelevant data may contain data from multiple irrelevant tasks or domains and can represent different combinations of such irrelevant sources. In addition, a set of irrelevant data may come from only a subset of the irrelevant domains.
\Cref{tab:scheme-T3,tab:scheme-T4,tab:scheme-T5,tab:scheme-T6,tab:scheme-T7} illustrate the high-level structure of these task distribution schemes.

\begin{table}[h]
\centering
\caption{Task distribution for \(|\mathcal{T}|=3\).}
\label{tab:scheme-T3}
\begin{tabular}{c|c|c}
\toprule
& \multicolumn{2}{c}{\(\mathcal{T}=\{t_1,t_2,t_3\}\)} \\
\cmidrule(lr){2-3}
Seller & Non-overlapping & Overlapping \\
\midrule
1 & \(\{t_1,t_2\}\) & \(\{t_1,t_2,t_3\}\) \\
2 & \(\{t_3\}\)     & \(\{t_1,t_2,t_4\}\) \\
3 & \(\{t_4\}\)     & \(\{t_1,t_4,t_5\}\) \\
4 & noisy/mislabeled/no data & \(\{t_4,t_5,t_6\}\) \\
\bottomrule
\end{tabular}
\end{table}

\begin{table}[h]
\centering
\caption{Task distribution for \(|\mathcal{T}|=4\).}
\label{tab:scheme-T4}
\begin{tabular}{c|cc|c}
\toprule
& \multicolumn{3}{c}{\(\mathcal{T}=\{t_1,t_2,t_3,t_4\}\)} \\
\cmidrule(lr){2-4}
Seller & \multicolumn{2}{c|}{Non-overlapping} & \multirow{2}{*}{\centering Overlapping} \\
\cmidrule(lr){2-3}
& Scheme A & Scheme B & \\
\midrule
1 & \(\{t_1,t_2,t_3\}\) & \(\{t_1,t_2\}\) & \(\{t_1,t_2,t_3,t_4\}\) \\
2 & \(\{t_4\}\)         & \(\{t_3\}\)     & \(\{t_1,t_2,t_3,t_5\}\) \\
3 & \(\{t_5\}\)         & \(\{t_5\}\)     & \(\{t_1,t_2,t_5,t_6\}\) \\
4 & noisy/mislabeled/no data
  & noisy/mislabeled/no data
  & \(\{t_1,t_5,t_6,t_7\}\) \\
5 & --- & --- & \(\{t_5,t_6,t_7,t_8\}\) \\
\bottomrule
\end{tabular}
\end{table}

\begin{table}[h]
\centering
\caption{Task distribution for \(|\mathcal{T}|=5\).}
\label{tab:scheme-T5}
\begin{tabular}{c|cc|c}
\toprule
& \multicolumn{3}{c}{\(\mathcal{T}=\{t_1,t_2,t_3,t_4,t_5\}\)} \\
\cmidrule(lr){2-4}
Seller & \multicolumn{2}{c|}{Non-overlapping} & \multirow{2}{*}{\centering Overlapping} \\
\cmidrule(lr){2-3}
& Scheme A & Scheme B & \\
\midrule
1 & \(\{t_1,t_2,t_3\}\) & \(\{t_1,t_2,t_3\}\) & \(\{t_1,t_2,t_3,t_4,t_5\}\) \\
2 & \(\{t_4,t_5\}\)         & \(\{t_4\}\)     & \(\{t_1,t_2,t_3,t_4,t_6\}\) \\
3 & \(\{t_6\}\)         & \(\{t_6\}\)     & \(\{t_1,t_2,t_3,t_6,t_7\}\) \\
4 & noisy/mislabeled/no data
  & noisy/mislabeled/no data
  & \(\{t_1,t_2,t_6,t_7,t_8\}\) \\
5 & --- & --- & \(\{t_1,t_6,t_7,t_8,t_9\}\) \\
6 & --- & --- & \(\{t_6,t_7,t_8,t_9,t_{10}\}\) \\
\bottomrule
\end{tabular}
\end{table}

\begin{table}[H]
\centering
\caption{Task distribution for \(|\mathcal{T}|=6\).}
\label{tab:scheme-T6}
\begin{tabular}{c|c|c}
\toprule
& \multicolumn{2}{c}{\(\mathcal{T}=\{t_1,t_2,t_3,t_4,t_5,t_6\}\)} \\
\cmidrule(lr){2-3}
Seller & Non-overlapping & Overlapping \\
\midrule
1 & \(\{t_1,t_2,t_3\}\) & \(\{t_1,t_2,t_3,t_4,t_5,t_6\}\) \\
2 & \(\{t_4,t_5\}\)     & \(\{t_1,t_2,t_3,t_4,t_5,t_7\}\) \\
3 & \(\{t_6\}\)     & \(\{t_1,t_2,t_3,t_4,t_7,t_8\}\) \\
4 & \(\{t_7\}\)     & \(\{t_1,t_2,t_3,t_7,t_8,t_9\}\) \\
5 & noisy/mislabeled/no data & \(\{t_1,t_2,t_7,t_8,t_9,t_{10}\}\) \\
6 & --- & \(\{t_1,t_7,t_8,t_9,t_{10},t_{11}\}\) \\
7 & --- & \(\{t_7,t_8,t_9,t_{10},t_{11},t_{12}\}\) \\
\bottomrule
\end{tabular}
\end{table}

\begin{table}[H]
\centering
\scriptsize
\caption{Task distribution for \(|\mathcal{T}|=7\).}
\label{tab:scheme-T7}
\begin{tabular}{c|ccc|c}
\toprule
& \multicolumn{4}{c}{\(\mathcal{T}=\{t_1,t_2,t_3,t_4,t_5,t_6,t_7\}\)} \\
\cmidrule(lr){2-5}
Seller & \multicolumn{3}{c|}{Non-overlapping} & \multirow{2}{*}{\centering Overlapping} \\
\cmidrule(lr){2-4}
& Scheme A & Scheme B & Scheme C & \\
\midrule
1 & \(\{t_1,t_2,t_3\}\) & \(\{t_1,t_2,t_3,t_4\}\) & \(\{t_1,t_2,t_3,t_4\}\) & \(\{t_1,t_2,t_3,t_4,t_5,t_6,t_7\}\) \\
2 & \(\{t_4,t_5\}\)     & \(\{t_5,t_6\}\)         & \(\{t_5,t_6,t_7\}\) & \(\{t_1,t_2,t_3,t_4,t_5,t_6,t_8\}\) \\
3 & \(\{t_6\}\)         & \(\{t_7\}\)             & \(\{t_8\}\) & \(\{t_1,t_2,t_3,t_4,t_5,t_8,t_9\}\) \\
4 & \(\{t_7\}\)         & \(\{t_8\}\)             & noisy/mislabeled/no data     & \(\{t_1,t_2,t_3,t_4,t_8,t_9,t_{10}\}\) \\
5 & noisy/mislabeled/no data
  & noisy/mislabeled/no data
  & ---
  & \(\{t_1,t_2,t_3,t_8,t_9,t_{10},t_{11}\}\) \\
6 & --- & --- & --- & \(\{t_1,t_2,t_8,t_9,t_{10},t_{11},t_{12}\}\) \\
7 & --- & --- & --- & \(\{t_1,t_8,t_9,t_{10},t_{11},t_{12},t_{13}\}\) \\
8 & --- & --- & --- & \(\{t_8,t_9,t_{10},t_{11},t_{12},t_{13},t_{14}\}\) \\
\bottomrule
\end{tabular}
\end{table}

\subsection{Impact Statement}\label{impact}
This work introduces DMVM, a framework for decentralized multi-task dataset valuation using model merging and secure multiparty computation. By enabling estimation of dataset utility without sharing raw data, our method supports fairer data marketplaces, incentivizes high-quality data contribution, and facilitates safer collaboration in sensitive domains such as healthcare and finance. At the same time, valuation mechanisms can be strategically manipulated or used to prioritize short-term performance over broader societal benefit. As with many advances in machine learning infrastructure, we encourage future work on governance, auditing, and fairness-aware valuation mechanisms to mitigate these risks.

\end{document}